\documentclass{article}

\usepackage[final]{neurips_2023}

\usepackage{appendix}
\usepackage[utf8]{inputenc} 
\usepackage[T1]{fontenc}    
\usepackage{tablefootnote}
\usepackage{hyperref}       
\usepackage{url}            
\usepackage{booktabs}       
\usepackage{amsfonts}       
\usepackage{nicefrac}       
\usepackage{microtype}      
\usepackage{xcolor}         
\usepackage{bm}
\usepackage{amsmath}
\usepackage{hyperref}
\usepackage{amssymb}
\usepackage{graphicx}
\usepackage{floatrow}
\usepackage{makecell}
\hypersetup{
    colorlinks=True,
    linkcolor=red,
    filecolor=green,      
    urlcolor=blue,
    citecolor=blue
}
\newfloatcommand{capbtabbox}{table}[][\FBwidth]

\usepackage{listings}

\definecolor{codegreen}{rgb}{0,0.6,0}
\definecolor{codegray}{rgb}{0.5,0.5,0.5}
\definecolor{codepurple}{rgb}{0.58,0,0.82}
\definecolor{backcolour}{rgb}{0.95,0.95,0.92}

\lstdefinestyle{mystyle}{
    backgroundcolor=\color{backcolour},   
    commentstyle=\color{codegreen},
    keywordstyle=\color{magenta},
    numberstyle=\tiny\color{codegray},
    stringstyle=\color{codepurple},
    basicstyle=\ttfamily\footnotesize,
    breakatwhitespace=false,         
    breaklines=true,                 
    captionpos=b,                    
    keepspaces=true,                 
    numbers=left,                    
    numbersep=2pt,                  
    showspaces=false,                
    showstringspaces=false,
    showtabs=false,                  
    tabsize=1
}

\title{Scaling MLPs: A Tale of Inductive Bias}

\author{Gregor Bachmann\thanks{Equal contribution. Correspondence to \texttt{\{gregorb, sanagnos\}}@ethz.ch. Code and checkpoints available \\
 \phantom{} \hspace{4mm} at \url{https://github.com/gregorbachmann/scaling_mlps}}\hspace{2mm}, \hspace{1mm} Sotiris Anagnostidis$^{*}$, \hspace{1mm} Thomas Hofmann \\[1mm]
ETH Z\"urich, Switzerland}

\begin{document}

\maketitle
\vspace{-3mm}
\begin{abstract}
In this work we revisit the most fundamental building block in deep learning, the multi-layer perceptron (MLP), and study the limits of its performance on vision tasks. Empirical insights into MLPs are important for multiple reasons. (1) Given the recent narrative \textit{"less inductive bias is better"}, popularized due to transformers eclipsing convolutional models, it is natural to explore the limits of this hypothesis. To that end, MLPs offer an ideal test bed, as they lack any vision-specific inductive bias. (2) MLPs have almost exclusively been the main protagonist in the deep learning theory literature due to their mathematical simplicity, serving as a proxy to explain empirical phenomena observed for more complex architectures. Surprisingly, experimental datapoints for MLPs are very difficult to find in the literature, especially when coupled with large pre-training protocols. This discrepancy between practice and theory is worrying: \textit{Do MLPs reflect the empirical advances exhibited by practical models?} Or do theorists need to rethink the role of MLPs as a proxy? We provide insights into both these aspects.
We show that the performance of MLPs drastically improves with scale ($95\%$ on CIFAR10, $82\%$ on CIFAR100, $58\%$ on ImageNet ReaL), highlighting that lack of inductive bias can indeed be compensated. We observe that MLPs mimic the behaviour of their modern counterparts faithfully, with some components in the learning setting however exhibiting stronger or unexpected behaviours. Due to their inherent computational efficiency, large pre-training experiments become more accessible for academic researchers. All of our experiments were run on a single GPU.
\vspace{-6mm}
\end{abstract}

\begin{figure}[!h]
    \centering
    \includegraphics[width=0.74\textwidth]{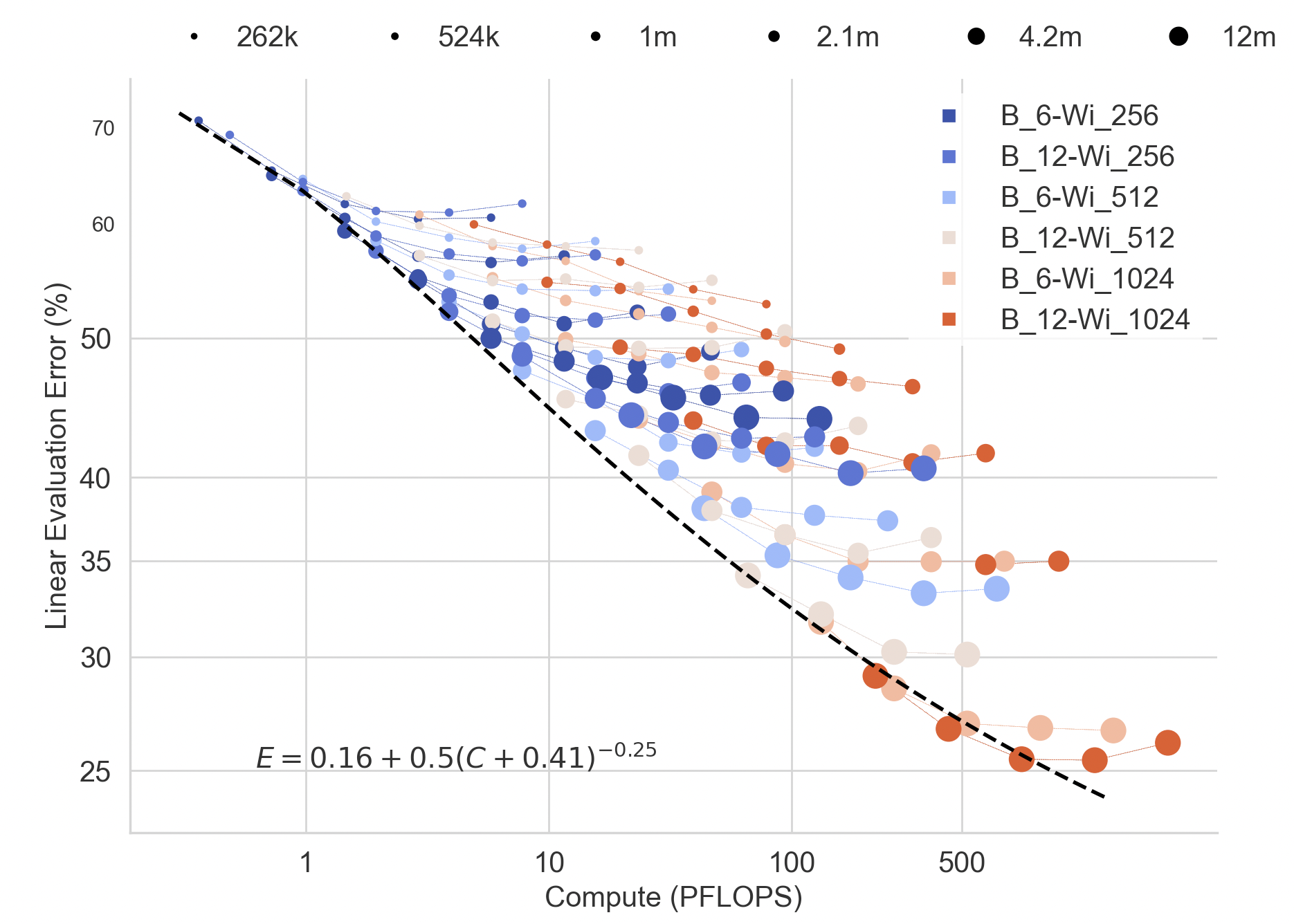}
    \caption{Test error on CIFAR100 as a function of PFLOPS.}
    \vspace{-3mm}
    \label{fig:scaling-cifar100}
\end{figure}

\section{Introduction}
Deep learning has undergone tremendous empirical progress in the last decades. The dominant approaches in practice these days rely on very large, pre-trained models which are then fine-tuned to the specific task at hand. For natural language processing, these models usually are some variant of the Transformer architecture \citep{NIPS2017_3f5ee243}, while in computer vision, both convolutional and transformer-based models are very popular \citep{He2015DeepRL, tan2020efficientnet,dosovitskiy2021an}. The theoretical understanding of these advances on the other hand remains very poor and the gap between the world of theory and practice is growing at an alarming rate. One aspect of this gap is the family of models investigated; due to their mathematical simplicity, theoretical works largely focus on simple multi-layer perceptrons (MLPs). Consisting of a series of unstructured matrix multiplications, interleaved with element-wise non-linearities, the MLP serves as an ideal test bed to analyze empirical phenomena exhibited by more complicated models employed in practice. Due to their inferior performance, MLPs are rarely used and very little is known regarding their behaviour in more modern settings. For instance, to the best of our knowledge, there is not a single published result showcasing an MLP trained on ImageNet1k, the de-facto standard benchmark in vision, let alone any pre-training/transfer learning studies. This lack of empirical data is concerning as theory aims to understand the characteristics of modern architectures through the lens of MLPs, yet only little assessments are made regarding how well such a proxy works. This raises the question,
\begin{equation}
    \vspace{2mm}
    \label{q:mlp-emp}
    \textit{Do MLPs reflect the empirical advances exhibited by practical models?}
    \vspace{2mm}
\end{equation}
Investigating MLPs is not only interesting for theory but also for practice. With the Vision Transformer (ViT) outperforming its convolutional competitors in very large-scale settings, the role of inductive bias has recently been brought into question. Since a ViT is equipped with significantly less inductive bias for vision  compared to convolutional models (e.g. it lacks translation-equivariance) a novel narrative has recently emerged:
\begin{equation}
    \vspace{2mm}
    \label{q:mlp-inductive}
    \textit{At large scales of compute, having less inductive bias is beneficial for performance.}  
    \vspace{2mm}
\end{equation}
More evidence for this hypothesis has been collected in the form of the MLP-Mixer \citep{tolstikhin2021mlpmixer}, an architecture with arguably even less inductive bias, solely relying on multi-layer perceptrons as patch processors and mixers. The MLP architecture is the ideal candidate to test the limits of such a hypothesis, as it exhibits the least inductive bias for vision due to its invariance to permutations of pixels. Unfortunately, the scale where Transformers and MLP-Mixers start to outperform convolutional models is out of reach for most researchers, requiring billions of annotated images and thousands of TPUs. We thus expect similar required scales for MLPs and hence instead investigate the following, weaker hypothesis:
\begin{equation}
    \vspace{2mm}
    \label{q:mlp-weaker}
    \textit{Lack of inductive bias can be compensated by scaling compute.} 
    \vspace{2mm}
\end{equation}
i.e. we aim to measure to what degree a lack of inductive bias hinders performance even if a model is subjected to a large parameter count and trained on datasets with many examples (albeit smaller than what is employed in \citet{dosovitskiy2021an}). \\[2mm]
\begin{figure}
    \centering \includegraphics[width=0.9\textwidth]{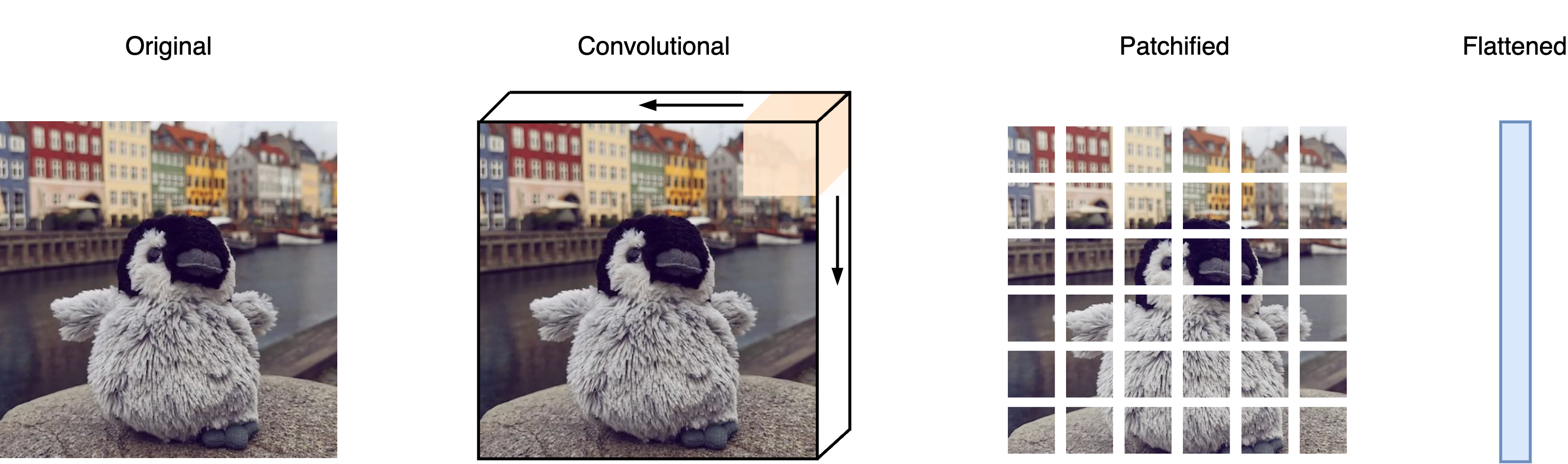}
    \caption{Different architectures process images differently. Convolutions directly operate on the image, ViTs and MLP-Mixers work with patches while the MLP takes the flattened image as input.}
    \label{fig:archs-img}
\end{figure}
In this work, we provide answers to question \ref{q:mlp-emp} and provide further evidence for hypothesis \ref{q:mlp-inductive} and \ref{q:mlp-weaker} by investigating how far we can push the empirical performance of models solely built from composing several MLP blocks. We give largely positive answers to question \ref{q:mlp-emp}, observing that MLPs behave very similarly to their modern counterparts when subjected to scale, i.e. their performance increases predictably as a power law in parameter count and sample size, akin to \citet{hestness2017deep,hestness2019humanlevel, kaplan2020scaling, Zhai_2022_CVPR} (see e.g. Fig.~\ref{fig:scaling-cifar100}). In contrast to previous work however, we find that compute-optimal MLPs allocate their budget more strongly into sample size, highlighting again their small inductive bias. While regularization in the form of data augmentation is also helpful for CNNs, its role is significantly amplified for MLPs even at large sample sizes, leading to fatal degradation if turned off. We further investigate how the implicit bias of SGD affects performance, and we make a very counter-intuitive discovery: contrary to CNNs, we find that larger batch sizes generalize significantly better for MLPs. This result questions the validity of the proxy role that the MLP plays in theoretical works investigating the implicit bias of SGD. While, as expected, the scale employed in this work does not suffice for hypothesis \ref{q:mlp-inductive}, we provide strong evidence for \ref{q:mlp-weaker}, which we view as an important first step. We observe that enough scale indeed suffices to overcome the bad inductive bias present in MLPs, leading to surprisingly strong downstream performance, e.g. $\approx 95\%$ on CIFAR10, $\approx 82\%$ on CIFAR100 and $\approx 58\%$ on ImageNet ReaL. In summary, we make the following contributions:
\begin{itemize}
    \item We fill the gap between theory and practice, providing the first results for MLPs trained in modern settings.
    \item We show that MLPs mostly behave comparably to their modern counterparts, making them a good proxy for theory. We observe however that the roles of regularization and implicit bias of SGD significantly differ and theory hence needs to adapt.
    \item We provide further evidence that inductive bias is not crucial at large scales, showing that even "bad" architectures like MLPs can achieve strong downstream performance. We however identify a shift in compute-optimality, showing that optimal MLPs invest their compute significantly more into dataset size compared to model size.
\end{itemize}

\section{Background}
\label{sec:background}
\paragraph{Theoretical Works.} The MLP has served as the main object of study for theoretical works in deep learning across different domains. The cornerstone results for areas such as convergence of SGD-trained neural networks \citep{doi:10.1073/pnas.1806579115, du2018gradient, zou2020gradient, NIPS2017_a96b65a7, saxe2014exact}, most generalization bounds \citep{Arora2019FineGrainedAO, mei2021generalization, NEURIPS2018_5a4be1fa, NEURIPS2019_62dad6e2}, the benefits of overparametrization \citep{neyshabur2018the, pmlr-v97-allen-zhu19a, pmlr-v80-arora18a}, the implicit bias of SGD towards favourable solutions \citep{soudry2018the, neyshanur2014implicit, pmlr-v125-chizat20a}, signal propagation properties \citep{NIPS2016_14851003, schoenholz2017deep} and scaling laws \citep{bahri2021explaining, maloney2022solvable} are all largely obtained for MLPs. To quote the very influential \textit{Principles of Deep Learning Theory} book \citep{roberts_yaida_hanin_2022}: \begin{center}
\textit{"MLPs are the simplest of these neural network architectures that hinge on this stacking
idea, and thus provide a minimal model for an effective theory of deep learning."}
\end{center}
There are also several theoretical works studying more modern setups such as convolutional or transformer-based networks including \citet{Arora2019OnEC, NEURIPS2018_0e98aeeb, pmlr-v70-brutzkus17a, pmlr-v119-hron20a} to name but a few, but the main theoretical focus to the best of our knowledge still remains on the MLP architecture. We thus believe it is important to explore the limits of such a theoretical proxy in realistic settings. 

\begin{figure}
    \centering
    \includegraphics[width=0.76\textwidth]{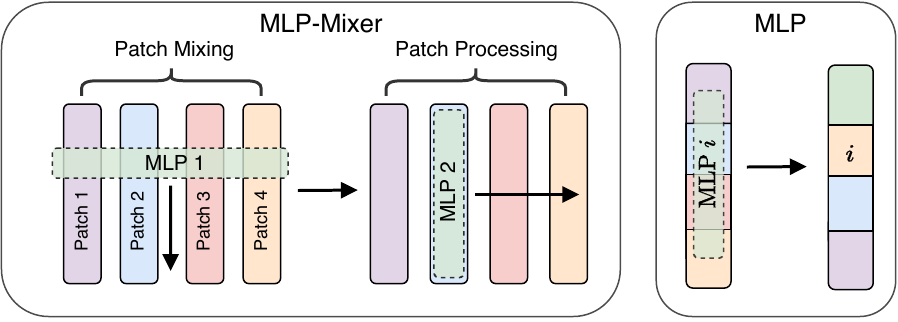}
    \caption{A simplified depiction of the differences between an MLP-Mixer and an MLP.}
    \label{fig:mlpmixer}
\end{figure}
\paragraph{MLPs.}
The multi-layer perceptron has its origins in \citet{rosenblatt1958perceptron}, serving as an extension to the classic Perceptron with its hidden layers however fixed to random initialization. \citet{ivakhnenko1965cybernetic} devised the first method to update the hidden layers through \textit{self-organization}. \citet{4039068} then introduced the idea to train the parameters with stochastic gradient descent. Mathematically, an MLP of depth $L \in \mathbb{N}$ can be described very efficiently; given an input $\bm{x} \in \mathbb{R}^{d}$, it applies a series of linear transformations, interleaved with an element-wise non-linearity $\sigma:\mathbb{R} \xrightarrow[]{}\mathbb{R}$:
\begin{equation*}
    \bm{z}^{(l)} = \bm{W}^{(l)}\bm{x}^{(l-1)} \hspace{2mm}\xrightarrow[]{\makebox[5mm]{}}\hspace{2mm}\bm{x}^{(l)} = \sigma\left(\bm{z}^{(l)}\right)
\end{equation*}
where we define $\bm{x}^{(0)} := \bm{x}$ and $\bm{W}^{(l)} \in \mathbb{R}^{d_l \times d_{l-1}}$ for $l=1, \dots, L$ are the learnable weight matrices. For the sake of readability, we omit the biases. This mathematical simplicity makes the MLP a very attractive model to study from a theoretical perspective (albeit still very far from trivial) and indeed many works frame their results around this more general model class. When used for vision, the input tensor $\bm{x} \in \mathbb{R}^{h \times w \times 3}$ is flattened into a vector $\operatorname{vec}(\bm{x}) \in \mathbb{R}^{3hw}$ and then passed through the MLP. Notice how such an architecture completely lacks locality and weight sharing, every unit simply processes the entire image at once. More worryingly, the vectorization $\operatorname{vec}$ could be applied in any way, i.e. any permutation of $\bm{x}$ looks identical to an MLP.\\[2mm] We want to highlight that MLPs of course are not completely free of inductive bias, in the sense that they encourage learning a hierarchical feature structure. On the other hand, there is no vision-specific inductive bias present in MLPs, which is the main setting we investigate here. We refer to \citet{battaglia2018relational} for a more in-depth treatment of inductive bias.

\paragraph{Convolutions.} The MLP is a very general model and has no structure built into it to make it more suitable for vision tasks. A convolution on the other hand was designed specifically for vision with desirable characteristics incorporated into the model. A convolution can be viewed as a special case of an MLP, where the weight matrix $\bm{W}$ is very structured by being sparse and having shared entries, leading to spatially localized learning. This can be most easily illustrated in the case of convolving a $2\times3 \times 1$ image $\bm{x}$ with a $2\times2$ filter $\bm{f}$ as the following matrix multiplication:
$$\bm{f} * \bm{x} = \bm{W}_{\bm{f}} \operatorname{vec}(\bm{x}) = \begin{pmatrix}
    f_1 & f_2  & 0 & f_3 & f_4 & 0  \\
    0 & f_1 & f_2 & 0 & f_3 & f_4 
\end{pmatrix} 
\operatorname{vec}(\bm{x})$$
Here $\operatorname{vec}$ denotes the standard, row-wise vectorization-scheme to flatten the image.
Instead of operating with a dense matrix as the MLP, the convolution uses a structured matrix $\bm{W}_{\bm{f}}$ tailored to the task of vision, leading to a better inductive bias. Moreover, a convolution exhibits translation-equivariance, i.e. shifts of images are processed equivalently to the original. Crucially, in contrast to the MLP, a convolution severely suffers if a permutation is applied to the image.

\paragraph{Vision Transformer.} Inspired by the successes in NLP, recently the Transformer architecture has been adapted to vision \citep{dosovitskiy2021an}. An image $\bm{x} \in \mathbb{R}^{h \times w \times 3}$ is broken up into smaller patches (also called tokens) and each such patch is linearly embedded (see Fig.~\ref{fig:archs-img}) and augmented with a so-called positional embedding, marking its spatial location in the image. The obtained embeddings are then processed by self-attention layers where patches can exchange information, and MLP layers, which are shared among patches and transform them individually. While the inductive bias of a ViT is certainly weaker compared to a CNN (it lacks translation-equivariance), the patching and parameter sharing still make the architecture suitable for vision.  

\paragraph{MLP-Mixer.} Similar to the ViT, the MLP-Mixer also works with a patchified image \citep{tolstikhin2021mlpmixer}. Unlike the ViT, token-mixing is not implemented using self-attention but rather another MLP block is used to exchange information between patches. We want to clearly highlight the difference between an MLP-Mixer and an MLP: An MLP-Mixer operates on patches, where in each block it applies a shared MLP to each patch for processing, and another MLP for mixing the patches along the channels. We visualize the differences in Fig.~\ref{fig:mlpmixer} for clarity. We again want to stress that breaking the image into patches and sharing parameters among them significantly enhances the amount of inductive bias, compared to a standard MLP.
\paragraph{Patchifiying.} As highlighted above, ViTs and Mixers largely obtain their inductive biases through breaking the images into patches. This choice seems to be beneficial even for architectures that already possess a strong inductive bias, such as the ConvMixer \citep{trockman2022patches}, where convolutions are performed on individual patches. The very recent Metaformer \citep{yu2022metaformer} further shows that even a simple spatial pooling instead of attention can lead to strong performance if the image is patchified. While the success of this mechanism certainly warrants further investigation, in this work we decided to deliberately focus on MLPs as they specifically lack this type of bias.
 
\begin{table}
\vskip 0.15in
\begin{center}
\begin{small}
\begin{sc}
\begin{tabular}{lccccc}
\toprule
 & CIFAR10 & CIFAR100  & TinyImageNet & ImageNet \\
\midrule
S-MLP (@$100$ E)  & $54.2$ & $28.8$ & $8.5$  & $9.2$ \\[1mm]
\midrule
S-MLP + DA (@ 1000 E) & $68.9$ &  $43.3$ & $25.2$ & $24.3$ \\[1mm]
\midrule
S-MLP + DA (@ 5000 E) & $72.3$ &  $44.5$ & $27.3$ & $26.8$ \\[1mm]
\midrule
B-MLP (@ 100 E)  & $58.1$  & $30.5$ & $8.9$ & $8.7$ \\[1mm]
\midrule
B-MLP + DA (@$1000$ E) & $70.1$ & $48.3$ & $27.2$  & $28.7$ \\[1mm]
\midrule
B-MLP + DA (@$5000$ E) & $75.4$ & $50.4$ & $31.2$  & $31.7$ \\[1mm]
\midrule 
ResNet18\tablefootnote{In contrast to the MLPs, the ResNet18 was trained at the original image resolutions.} + DA & $93.2$ & $75.6$ & $68.9$ & $69.7$ \\[1mm]
\bottomrule
\end{tabular}
\end{sc}
\end{small}
\end{center}
\caption{Test accuracies (in $\%$) without any pre-training. The S-MLP has depth $6$ and width $1024$ while the B-MLP has depth $6$, width $1024$ and an expansion factor of $4$. }
\label{tab:accs-scratch}
\end{table}

\section{Architecture}
We study different variants of the MLP architecture, starting from the standard vanilla setup and then adding more components such as residual connections and bottleneck layers. 
\paragraph{Standard MLP.} As a first starting point, we investigate simple MLPs with ReLU activations and isotropic design, i.e. except for the first, every layer has the same width $m \in \mathbb{N}$. In order to avoid training instabilities we further enhance the standard MLP with layer normalizations \citep{ba2016layer} placed after the activations. We thus compose several blocks of the form
$$\operatorname{Block}(\bm{z}) = \sigma\left(\bm{W}\operatorname{LN}(\bm{z})\right)$$
with $\bm{W} \in \mathbb{R}^{m \times m}$. To embed the image $\bm{x} \in \mathbb{R}^{d \times d \times 3}$ we use a linear layer $\operatorname{emb}(\bm{x}) = \bm{W}^{emb} \operatorname{vec}(\bm{x})$ with $\bm{W}^{emb} \in \mathbb{R}^{m \times 3d^2}$. Such an embedding layer is crucial since for high resolution images, $3d^2$ can be quite large and thus $m$ needs to be chosen smaller.
We empirically find that such a network design is the minimal choice in order to guarantee successful training across all scales of parameter count and sample size. We will use the short cut \textit{S-MLP} to denote such an architecture.
\paragraph{Inverted Bottleneck MLP.} Inspired by \citet{Lin2015HowFC, tolstikhin2021mlpmixer} we add a bottleneck structure to an MLP block as well as skip connections as follows:
$$\operatorname{Block}(\bm{z}) = \bm{z} + \bm{W}^{c}\sigma\left(\bm{W}^{e}\operatorname{LN}\left(\bm{z}\right)\right)$$
where $\bm{W}^{e} \in \mathbb{R}^{km \times m}$ expands the dimension to $km$ for $k \in \mathbb{N}$ and $\bm{W}^{(c)} \in \mathbb{R}^{m \times km}$ collapses it back to width $m$. For most experiments we set $k=4$. While the additions of skip connections and bottleneck layers to the architecture arguably add some amount of inductive bias, we  believe that in comparison to modern architectures such enhancements remain negligible. We will denote this variant by \textit{B-MLP}.
\section{Experiments}
\subsection{Setup}
In this work, we solely focus on vision tasks as inductive bias is more readily understood in this setting. Moreover, most theoretical works focus on image classification tasks, making it thus a natural test bed to assess the performance of MLPs. We study the popular tasks CIFAR10, CIFAR100 \citep{Krizhevsky2009LearningML}, STL10 \citep{pmlr-v15-coates11a}, TinyImageNet \citep{Le2015TinyIV}, ImageNet1k for evaluation,  as well as ImageNet21k \citep{5206848} for pre-training. In order to limit the size of the embedding layer and the computational needs, we downscale all images to resolution $64 \times 64 \times 3$ (if needed) as done in \citet{chrabaszcz2017downsampled}. We center and normalize all the images as a pre-processing step. For data augmentations, we consider random flips and crops as well as MixUp \citep{zhang2018mixup}.
\begin{table}
\vskip 0.15in
\begin{center}
\begin{small}
\begin{sc}
\begin{tabular}{lcccccc}
\toprule
 & CIFAR10 & CIFAR100 & STL10 & Tiny-IN & IN & ReaL \\
\midrule
\textit{{B}-6/{Wi}-1024}  & $69.9$\kern-.1ex\raisebox{.15ex}{\tiny{${\pm 0.1}$}}\kern-.40ex & $43.0$\kern-.1ex\raisebox{.15ex}{\tiny{${\pm 0.4}$}}\kern-.40ex & $51.5$\kern-.1ex\raisebox{.15ex}{\tiny{${\pm 0.1}$}}\kern-.40ex  & $47.1$\kern-.1ex\raisebox{.15ex}{\tiny{${\pm 0.1}$}}\kern-.40ex & $15.2$\kern-.1ex\raisebox{.15ex}{\tiny{${\pm 0.2}$}}\kern-.40ex &
$20.3$\kern-.1ex\raisebox{.15ex}{\tiny{${\pm 0.2}$}}\kern-.40ex
\\[1mm]
\midrule
\textit{{B}-6/{Wi}-1024} + DA  & $91.5$\kern-.1ex\raisebox{.15ex}{\tiny{${\pm 0.02}$}}\kern-.40ex & $76.4$\kern-.1ex\raisebox{.15ex}{\tiny{${\pm 0.2}$}}\kern-.40ex & $85.0$\kern-.1ex\raisebox{.15ex}{\tiny{${\pm 0.2}$}}\kern-.40ex  & $62.7$\kern-.1ex\raisebox{.15ex}{\tiny{${\pm 0.1}$}}\kern-.40ex & $38.7$\kern-.1ex\raisebox{.15ex}{\tiny{${\pm 0.1}$}}\kern-.40ex &
$47.0$\kern-.1ex\raisebox{.15ex}{\tiny{${\pm 0.15}$}}\kern-.40ex\\[1mm]
\midrule
\textit{{B}-12/{Wi}-1024} + DA & $94.2$\kern-.1ex\raisebox{.15ex}{\tiny{${\pm 0.05}$}}\kern-.40ex & $80.0$\kern-.1ex\raisebox{.15ex}{\tiny{${\pm 0.05}$}}\kern-.40ex & $89.9$\kern-.1ex\raisebox{.15ex}{\tiny{${\pm 0.1}$}}\kern-.40ex & $69.9$\kern-.1ex\raisebox{.15ex}{\tiny{${\pm 0.4}$}}\kern-.40ex & $43.3$\kern-.1ex\raisebox{.15ex}{\tiny{${\pm 0.06}$}}\kern-.40ex &
$48.6$\kern-.1ex\raisebox{.15ex}{\tiny{${\pm 0.2}$}}\kern-.40ex\\[1mm]
\midrule
\textit{{B}-12/{Wi}-1024} + DA + TTA & $95.5$\kern-.1ex\raisebox{.15ex}{\tiny{${\pm 0.05}$}}\kern-.40ex & $82.6$\kern-.1ex\raisebox{.15ex}{\tiny{${\pm 0.2}$}}\kern-.40ex & $92.2$\kern-.1ex\raisebox{.15ex}{\tiny{${\pm 0.05}$}}\kern-.40ex & $73.1$\kern-.1ex\raisebox{.15ex}{\tiny{${\pm 0.5}$}}\kern-.40ex & $51.5$\kern-.1ex\raisebox{.15ex}{\tiny{${\pm 0.1}$}}\kern-.40ex &  $57.9$\kern-.1ex\raisebox{.15ex}{\tiny{${\pm 0.1}$}}\kern-.40ex\\[1mm]

\bottomrule
\end{tabular}
\end{sc}
\end{small}
\end{center}
\caption{Fine-tuning Top-1 accuracies (in $\%$) when pretrained on ImageNet21k. Accuracies are averaged over $3$ runs. For readability, we abbreviate ImageNet as IN.}
\label{tab:finetune}
\end{table}
\subsection{Training from Scratch}
We start the empirical exploration of MLPs by training them from scratch (i.e. without any extra data) on popular vision benchmarks. All models were trained with the LION optimizer \citep{chen2023symbolic} with a learning rate $\eta=5\mathrm{e}$-$5$. In order to combat overfitting we use strong label smoothing $\alpha=0.3$.  We display the resulting test accuracies in Table~\ref{tab:accs-scratch}. We observe that both the standard architecture and the bottleneck without any data augmentation suffer from overfitting, leading to suboptimal performance. When turning it on, data augmentation as a regularizer however really unfolds its full power, significantly pushing the performance by roughly $20\%$ across all tasks. As observed in \citet{Lin2015HowFC}, the inverted bottleneck architecture leads to an improvement in performance across all datasets. Learning on the other hand significantly slows down with strong augmentations such as MixUp, enabling training for up to $5000$ epochs without suffering from overfitting. However, compared to simple modern baselines such as a ResNet18 \citep{He2015DeepRL}, a large discrepancy in performance remains, highlighting the importance of inductive bias in the small sample regime. We remark that ViTs and MLP-Mixers as well exhibit more learning difficulties if the dataset size is small \citep{dosovitskiy2021an, tolstikhin2021mlpmixer}. We provide more ablation studies in Appendix \ref{sec:ablations}.

\subsection{Transfer Learning}
In this section, we aim to analyze how transferable features learnt by MLPs are across different vision tasks. Transferability is one of the hallmark characteristics of modern deep learning, enabling practitioners to fine-tune large models on their specific dataset, leading to superior performance. We are, to the best of our knowledge, the first to measure transferability of MLPs, which is crucial to assess in order to build a theoretical understanding of the process. In this section, we focus on the inverted bottleneck MLP as it generalizes better and is easier to optimize. We provide the dual results for the standard MLP in Appendix \ref{sec:transfer-standard}.  We restrict to $k=4$ for the expansion factor and denote by $\textit{{B}-L/{Wi}-m}$ a network with $L$ blocks and width $m$. For pre-training we use ImageNet21k, the largest publicly available image dataset with annotated classes. After preprocessing the dataset following \citet{imagenetformasses}, it consists of roughly 12 million images and 11 thousand classes. We then pre-train the MLP with the cross-entropy loss for $800$ epochs, employing label smoothing and the LION optimizer. To guarantee fast data loading we rely on the FFCV framework \citep{leclerc2023ffcv} for all experiments. \\[3mm] 
In order to measure transferability of the learnt features we fine-tune the network on the new task. We also study training a linear layer on top of the embeddings but defer those results to Appendix \ref{sec:linear-probing}. We again explore the effects of data augmentation during the pre-training stage. For fine-tuning we use SGD with momentum with a learning rate of $\eta_{\text{head}}=0.01$ for the head and $\eta_{\text{body}}=0.001$ for the encoder for $50$ epochs. We upscale CIFAR images to resolution $64 \times 64 \times 3$ at fine-tuning time to guarantee compatibility. We display the fine-tuning results in Table \ref{tab:finetune}. For visualizations of the learnt features, we refer the interested reader to Appendix \ref{sec:visualizations-features}. We again observe that using data augmentation during the pre-training phase is essential to successful training, boosting performance up to $30\%$ in case of CIFAR100. Surprisingly, the learnt features are highly transferable, improving the performances reported previously in Table \ref{tab:accs-scratch} dramatically. While of course pre-trained on a large quantity of data, we nevertheless want to highlight that such an MLP becomes competitive with a ResNet18 trained from scratch for all the datasets, except for ImageNet1k where performance falls surprisingly short. We hypothesize that MLPs struggle with the more fine-grained distinctions between classes, in combination with the reduced resolution of the images.

\paragraph{Test-Time Augmentations.}  For ImageNet1k we further notice that objects tend to not be centered, in contrast to datasets like CIFAR10. We suspect that this might lead to the comparatively weaker performance. To test this, we leverage test-time augmentations (TTA). As introduced by \citet{NIPS2012_c399862d}, for each test image, we produce a fixed number of $100$ random crops and use the averaged logits for prediction. We observe significant improvements across all datasets, especially for ImageNet we obtain an increase of roughly $8\%$. This indeed indicates that MLPs struggle to localize the object of interest, especially for the more complicated ImageNet1k task. Using a large number of crops alleviates this problem to some degree. This also explains why the gains on tasks like CIFAR10 are smaller as the objects there usually are perfectly centered. 
\paragraph{ReaL accuary.} As observed in \citep{beyer2020imagenet}, the ImageNet labels do not capture that a single image might contain multiple objects of distinct classes. ImageNet accuracy can thus be misleading in the sense that model classes such as convolutional networks might have implicitly adapted to the particular labeling strategy due to the repeated benchmarking on the same validation set. MLPs most likely lack such an implicit adaptation as this work is to our knowledge the first to evaluate them on ImageNet1k. To address this, \citet{beyer2020imagenet} introduced a novel set of validation labels that better capture the multi-label nature, where a prediction is deemed correct if it matches one of the categories present in the image. We observe further very significant improvements of $\approx 7\%$ when employing ImageNet ReaL.  \\[2mm] 
Overall, these results underline that a bad inductive bias as exhibited by an MLP can indeed be overcome if subjected to enough scale. For theory, the results are double-edged; while MLPs prove to be a good proxy to understand transfer learning, data augmentation proves to be a crucial component. Also test-time augmentations significantly boost performance. Both these components on the other hand remain rather understudied in theoretical works.

\begin{figure}
\begin{floatrow}
\ffigbox{%
  \includegraphics[width=0.44\textwidth]{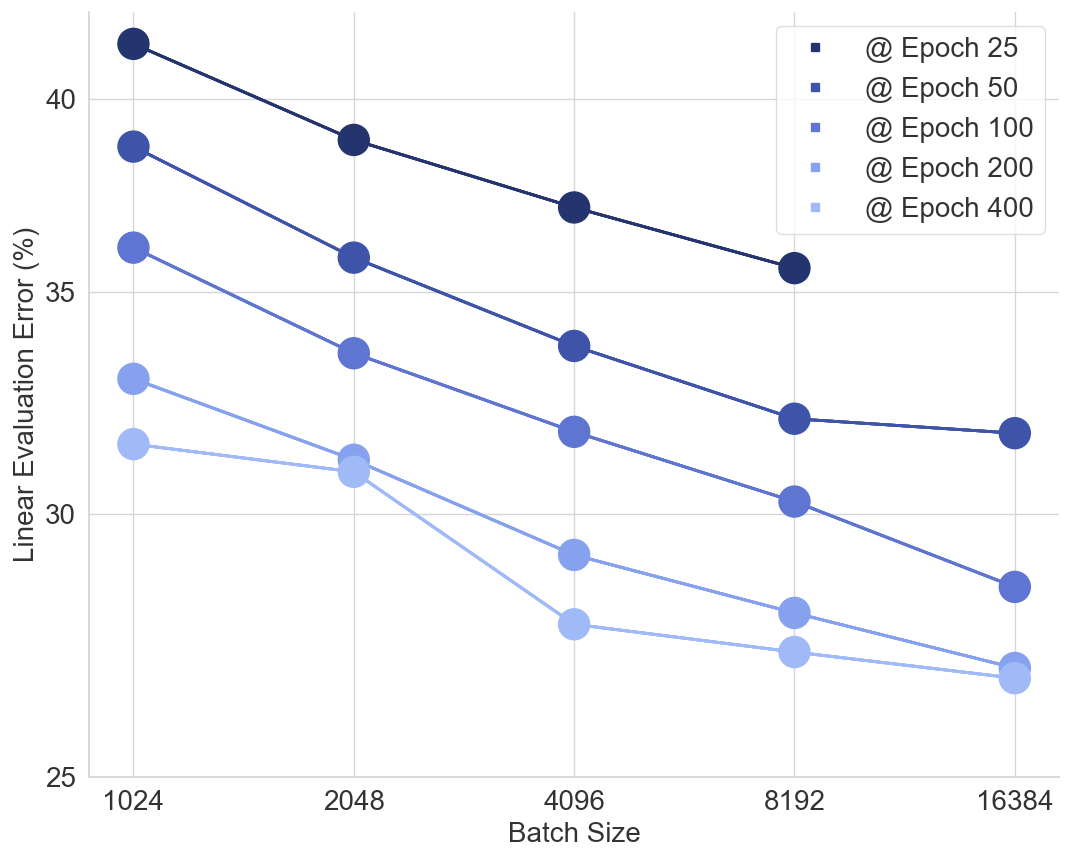}
}{%
  \caption{Linear downstream error on CIFAR100 (in $\%$) when pretrained for varying batch-sizes on ImageNet21k, on a log-log scale.}
  \label{fig:batch_size}
}

\capbtabbox{%
  \begin{tabular}{lc}
\toprule
Model & $\#$parameters  \\
\midrule
\textit{{B}-6/{Wi}-256}  & $9$M  \\[1mm]
\midrule
\textit{{B}-12/{Wi}-256}   & $12$M  \\[1mm]
\midrule
\textit{{B}-6/{Wi}-512}   & $24$M  \\[1mm]
\midrule
\textit{{B}-12/{Wi}-512}   & $37$M  \\[1mm]
\midrule
\textit{{B}-6/{Wi}-1024}   & $74$M  \\[1mm]
\midrule
\textit{{B}-12/{Wi}-1024} & $124$M  \\[1mm]

\bottomrule
\end{tabular}
}{%
  \caption{The different models and the respective parameter counts in millions.}%
  \label{tab:parameter-count}
}
\end{floatrow}
\end{figure}

\paragraph{Large batch-sizes.}
We further make the counter-intuitive observation that training with larger batch sizes significantly boosts performance both up- and downstream. In Fig.~\ref{fig:batch_size} we plot pre-training batch size against resulting linear downstream accuracy on CIFAR100 for different number of pre-training epochs. We observe that across all training times, using a larger batch size leads to significantly better performance. Moreover, we want to highlight that such a plot is even favoring small batch-sizes since those models perform more gradient updates for a fixed number of epochs. This effect is in stark contrast to convolutional architectures where entire lines of works have focused on preserving the performance of the small batch-size regime for larger ones \citep{imagenet1hour, you2017large, NIPS2017_a5e0ff62, keskar2017on}. Training with large batch-sizes without degradation is of high interest as it can lead to potentially more efficient training pipelines since computation can be sharded among more devices. This observation about optimal batch-sizes is in-line with similar recent conclusions in Transformers~\citep{kaplan2020scaling, touvron2023llama}.

\paragraph{Role of augmentations.} The role of data augmentation is very pronounced for MLPs, largely since it provides indirect inductive bias to the model. Remarkably, a model pre-trained on $12$ million examples without data augmentation shows inferior performance on CIFAR10 compared to a network trained from scratch with augmentations turned on. This emphasizes that augmentations go beyond merely leading to a bigger dataset but provide the model with useful invariances. We investigate the learnt weights in-depth in Appendix~\ref{sec:visualizations-features}, showing that very evidently, more localized features are learnt if data augmentation is employed. The power of augmentations has already been demonstrated previously through the advent of self-supervised learning \citep{NEURIPS2020_f3ada80d, Caron2021EmergingPI, pmlr-v119-chen20j}. Even when training on purely random labels, it still provides a powerful learning signal \citep{anagnostidis2023the}.

\subsection{Scaling Laws}
\label{sec:scaling-laws}
One of the key mysteries in deep learning is that networks tend to improve in terms of generalization when compute, in the form of parameter count and dataset size, is scaled up. Recently it has been observed in several works that the benefits of scale are highly predictable, i.e. generalization performance exhibits a power-law structure when plotted against compute measured in FLOPS \citep{rosenfeld2020a, hestness2017deep,hestness2019humanlevel, kaplan2020scaling, Zhai_2022_CVPR}. The functional form has recently been further refined \citep{caballero2023broken}.  The predictable nature of test performance has even been leveraged to estimate the optimal model before training \citep{hoffmann2022training, openai2023gpt4}. In order to understand this important characteristic of deep learning theoretically, it is important to analyze whether MLPs exhibit similar properties. 

\begin{figure}[h!]
    \centering
    \includegraphics[width=0.495\textwidth]{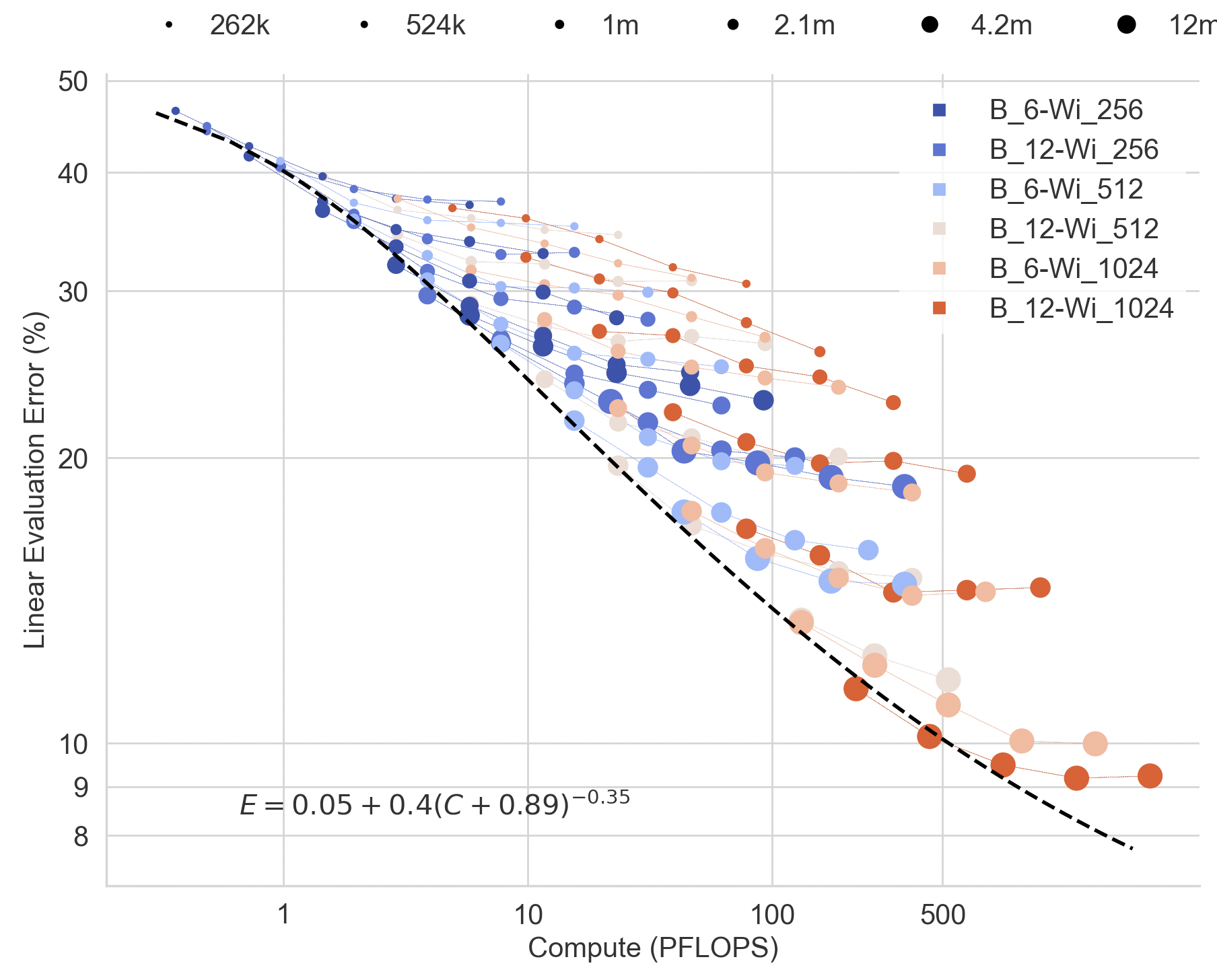}%
    \includegraphics[width=0.505\textwidth]{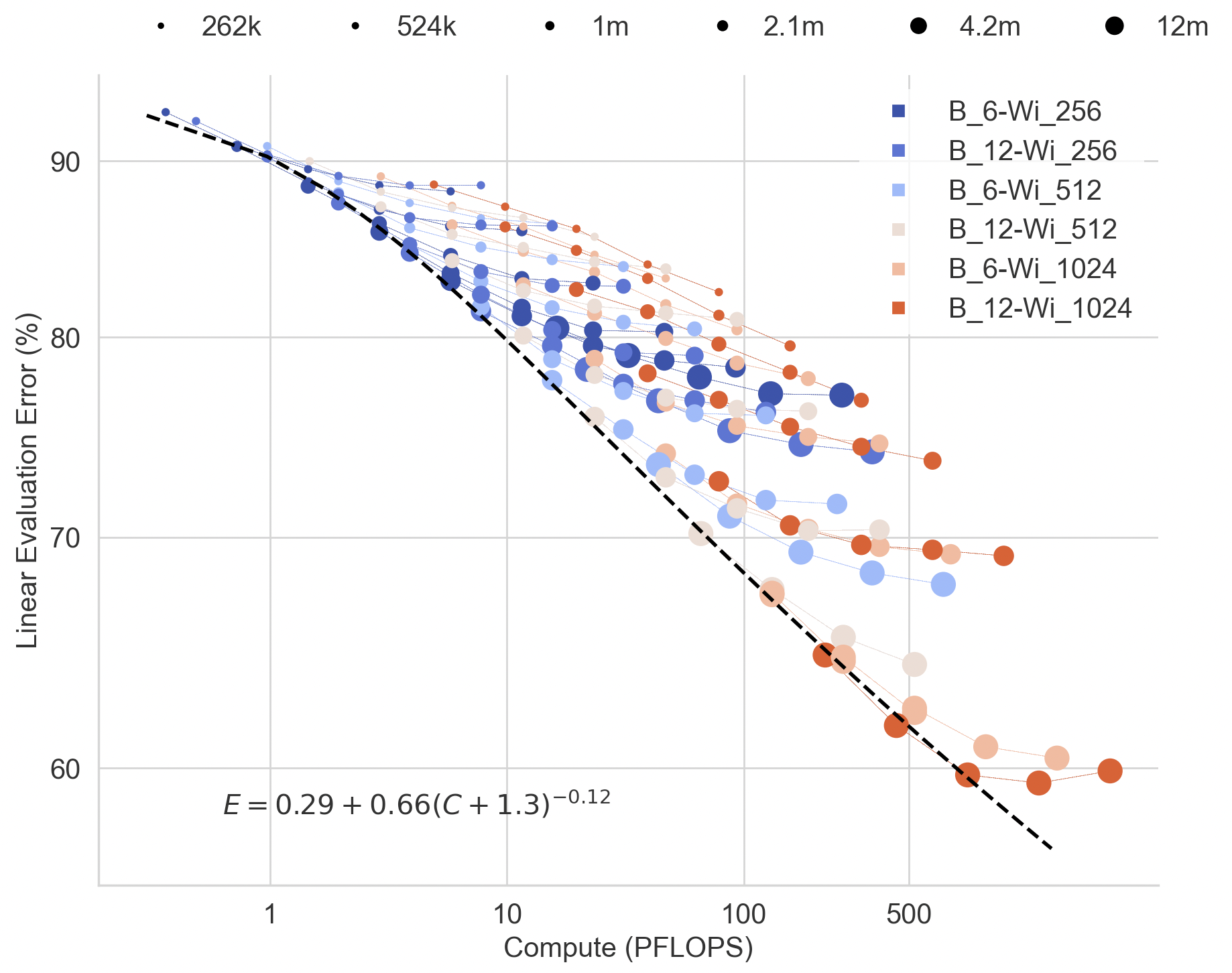}
    \caption{Test error (in $\%$) on CIFAR10 (left) and ImageNet1k (right) when linearly transferred as a function of PFLOPS, measured according to Eq.\eqref{eq:compute}, on a log-log scale.}
    \label{fig:cifar10-scaling}
\end{figure}

\begin{figure}
    \centering
\includegraphics[width=0.45\textwidth]{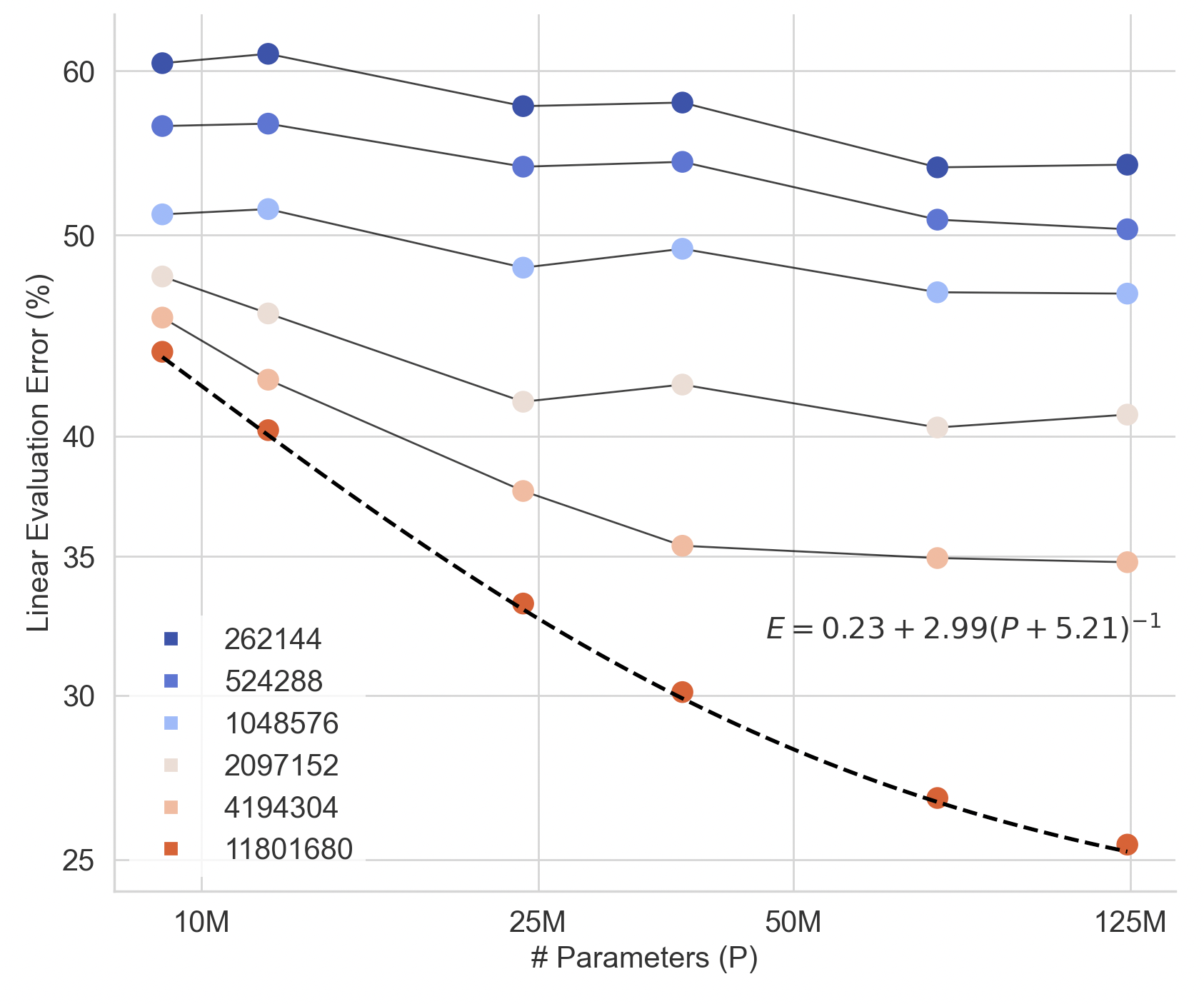}%
\includegraphics[width=0.49\textwidth]{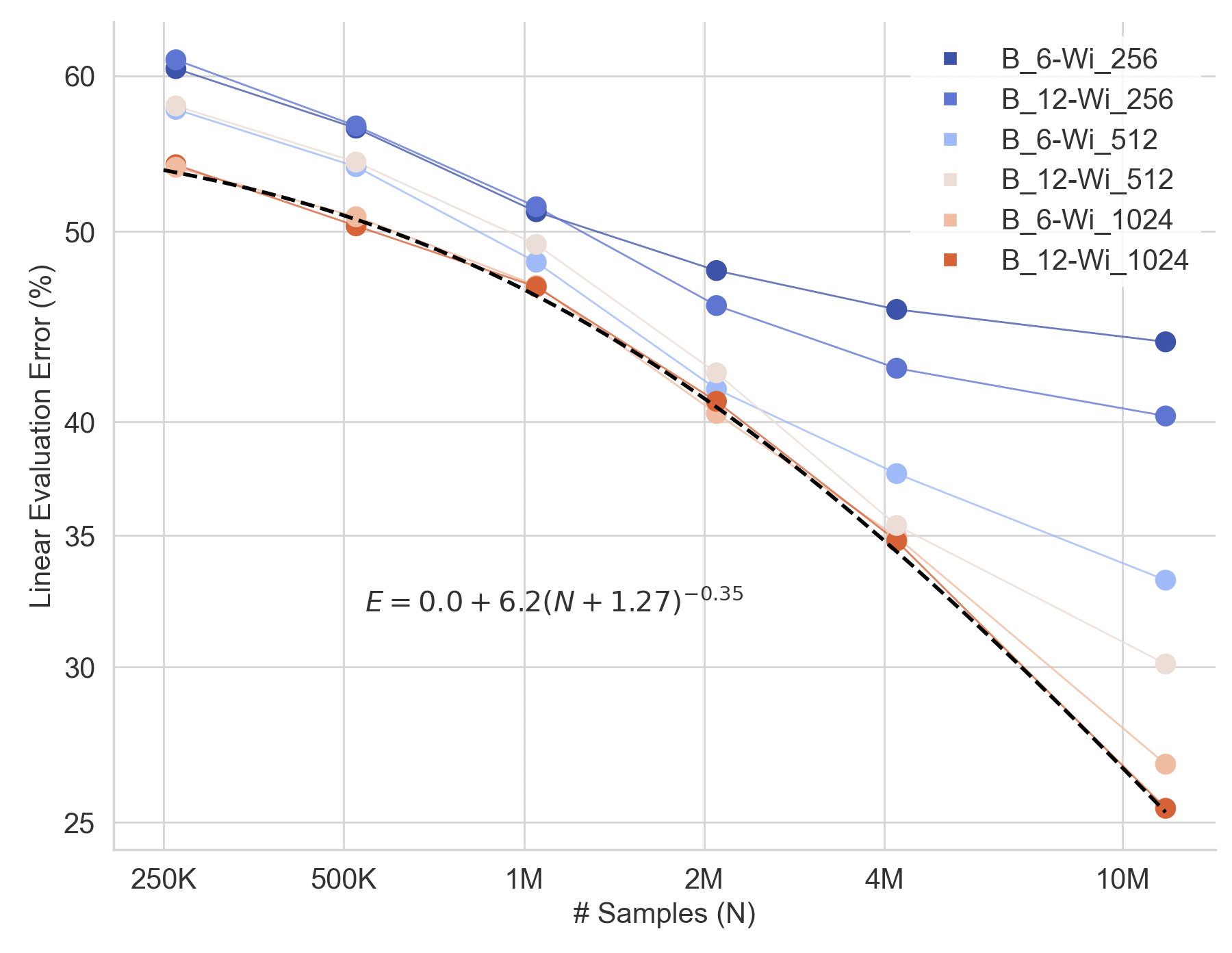}
    \caption{Power law in linear evaluation error on CIFAR100 (in $\%$) when either bottlenecked by the number of parameters (left) or the number of examples (right), on a log-log scale. The dotted line visualizes the fitted functional form.}
    \label{fig:ps-scaling-cifar100}
\end{figure}

\paragraph{Compute.} Following \citet{OPENAI_2018} we define the computational cost $C$ incurred from training a model $f$ on $N$ examples for $T$ epochs as 
\begin{equation}
    \label{eq:compute}
    C = \text{FLOP}(f) \times 3 \times N \times T,
\end{equation}
where $\text{FLOP}(f)$ denotes the number of FLOPs needed to complete the forward pass of $f$ for a single example. We note that the number of parameters $P$ present in $f$ enters this equation implicitly in the form of $\text{FLOP}(f) \propto P$. Observe that a given level of compute can be achieved in different ways, i.e. using more parameters $P$, training on more examples $N$, or training for a longer time $T$. When allocating a given level of compute optimally, it is observed that for convolutional and transformer-based architectures, the test error $E(C)$ as a function of compute behaves as a power-law
\begin{equation}
    \label{eq:powerlaw}
    E(C) = a(b + C)^{-\alpha}+E_{\infty},
\end{equation}
where $a,b,E_{\infty} \in \mathbb{R}_{+}$ and $\alpha>0$ is the scaling coefficient determining the rate of decay. $E_{\infty}$ denotes the irreducible error, i.e. even if infinite compute were employed, the performance remains imperfect. The test error can be measured upstream (i.e. on the pre-training task) or downstream when fine-tuning on a different task. We investigate various pre-training schemes with different number of examples, parameter counts and training times. We subsample ImageNet21k proportionally across classes and pre-train variously sized inverted bottleneck MLPs. We summarize the configurations in Table \ref{tab:parameter-count}. We then measure test error on the downstream task of CIFAR100 in Fig.~\ref{fig:scaling-cifar100} as well as CIFAR10 and ImageNet1k in Fig.~\ref{fig:cifar10-scaling} by linearly transferring the learnt features (without test-time augmentations). The plotting style is inspired by \citet{Zhai_2022_CVPR}. Each point in the curve is the downstream performance of an MLP, where the color of the point indicates the model type (blue denotes smaller and red larger models) and the size of the point indicates the number of pre-training examples. Points connected by a line indicates longer training times where $T \in \{50, 100, 200, 400, 800\}$ is measured in epochs. In all experiments, we employ data augmentation for pre-training. We observe that the compute-optimal performance of MLPs strongly exhibits characteristics of a power-law with coefficients $\alpha \in \{0.12, 0.25, 0.35\}$. This is very encouraging for future theoretical work, showing that MLPs indeed mirror the scaling behaviour of modern models. We provide the dual results for the standard MLPs in Appendix \ref{sec:scaling-standard}, noting that they exhibit essentially the same scaling behaviour, albeit with a slightly weaker slope and intercept. \\[2mm]
We further study how performance $E$ evolves when compute is either bottlenecked by the number of parameters $P$ or the dataset size $N$. We visualize the resulting scaling laws in Fig.~\ref{fig:ps-scaling-cifar100}. We find a very steep decay rate in terms of parameters $P$ where roughly $\alpha_P \approx 1$, whereas for dataset size $N$ we identify a significantly slower rate of $\alpha_N \approx 0.35$. This shows that the performance of MLPs is significantly more limited by the dataset size, which is in-line with the fact that MLPs exhibit a bad inductive bias. We investigate the role of dataset size and parameters more in the next paragraph.

\begin{figure}
    \centering
    \includegraphics[width=0.47\textwidth]{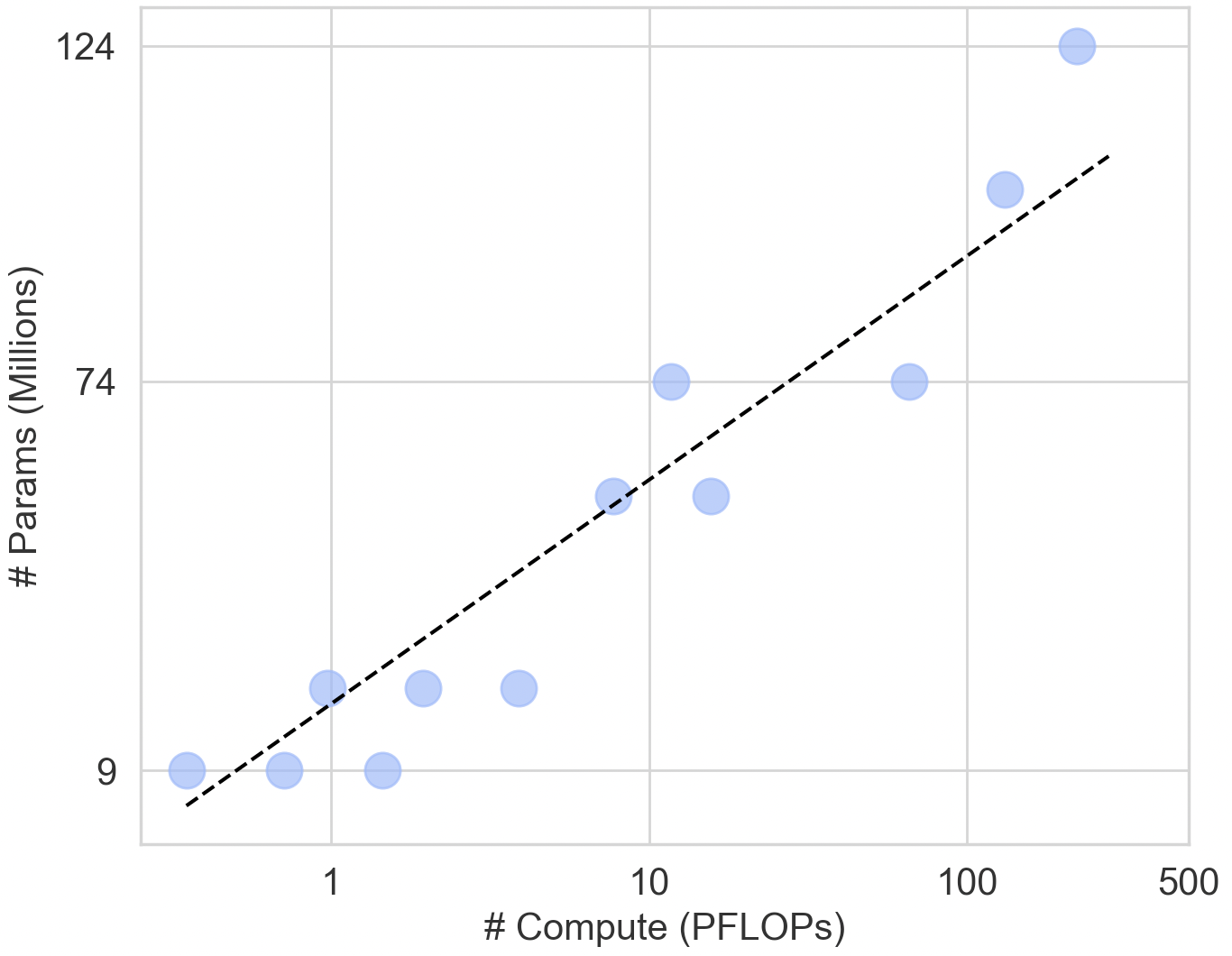}%
     \includegraphics[width=0.49\textwidth]{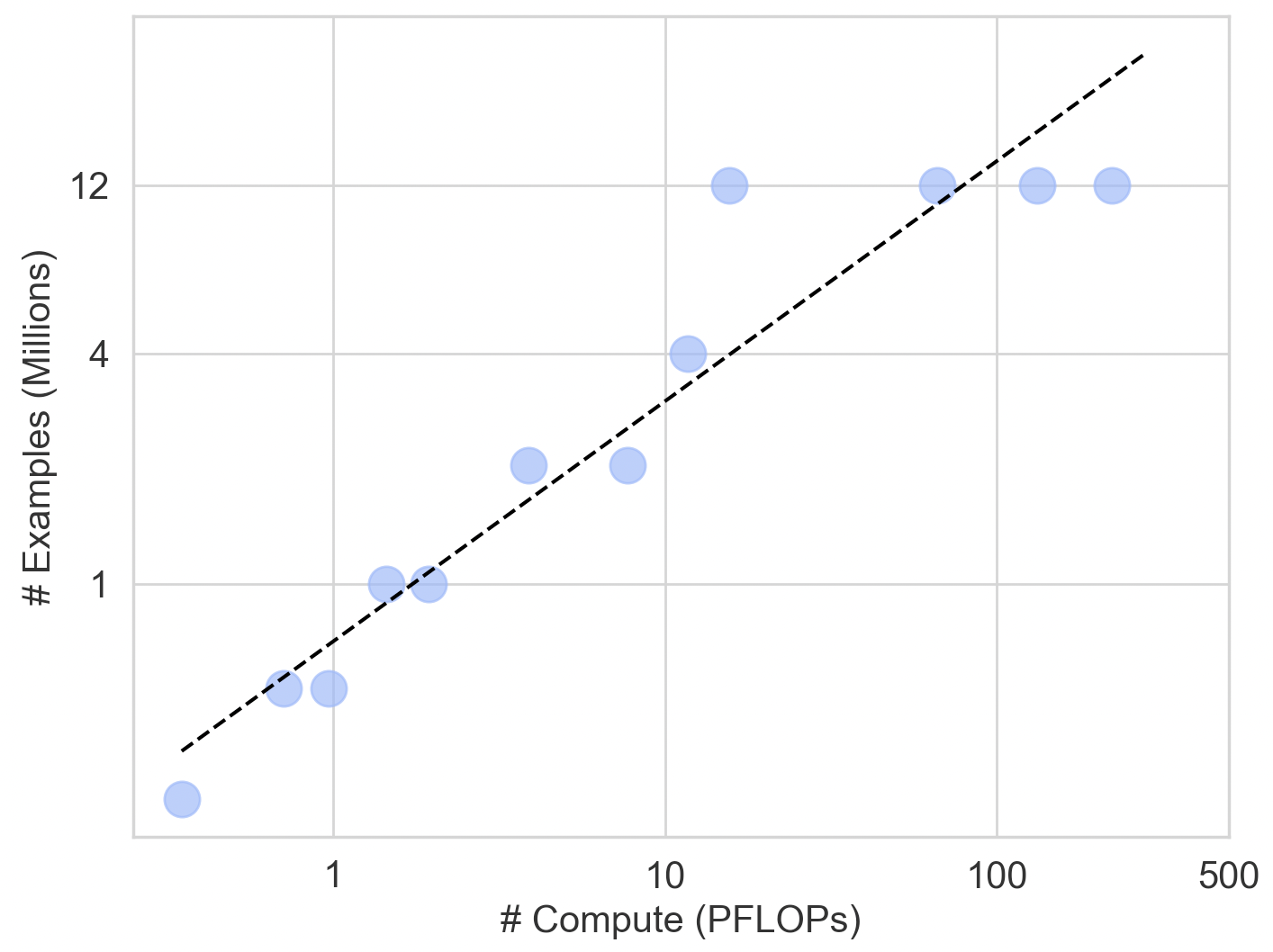}
    \caption{Optimal model size (left) and number of examples (right) for a given level of compute for linear evaluation on CIFAR100, on a log-log scale.}
    \label{fig:optimal-allocation}
\end{figure}

\paragraph{Parameters or examples.} Given a fixed level of compute $C$, what is the optimal way to allocate it to parameter count $P$ and number of examples $N$? In order to be more comparable to previous work, we assume a fixed training time $T=50$. To answer this question, we follow the approach outlined in \citet{hoffmann2022training} and plot the optimal compute models identified in Fig.~\ref{fig:scaling-cifar100} both against model size $P$ and number of examples $N$. We visualize the results in Fig.~\ref{fig:optimal-allocation}. We empirically observe that the optimal parameter count $P^{*}(C)$ and dataset size $N^{*}(C)$ as a function of compute $C$ exhibit power-law behaviour of the approximate form

$$P^{*}(C) \propto C^{0.35} \hspace{5mm} N^{*}(C) \propto C^{0.65}$$
While for transformers, the number of examples (or tokens) $N$ and parameters $P$ are scaled equally \citep{hoffmann2022training} (i.e. $\alpha_P \approx \alpha_N \approx 0.5$), in contrast we observe that the optimal strategy for MLPs invests significantly more compute into dataset size $N$. This is further evidence for the weaker inductive bias present in MLPs, which needs more examples in order to be compensated for. 


\subsection{Computational Feasibility}
We believe that a further exciting feature of our study is its computational feasibility, while at the same time preserving the main characteristics of large-scale pre-training. All of our experiments were conducted on a single NVIDIA RTX A5000 GPU with 24GB of memory. In conjunction with the strongly optimized FFCV dataloading framework \citep{leclerc2023ffcv} and the inherent efficiency of MLPs, we are able to perform very rapid training. For instance we complete a single epoch on ImageNet21k with the \textit{{B}-12/{Wi}-1024} architecture, equipped with $124$ million parameters, in only roughly $450$ seconds, while the smaller variant \textit{{B}-6/{Wi}-1024} at a parameter count of $74$ million requires roughly $250$ seconds on the specified hardware. Low memory requirements allow us to train with a batch-size of $16384$ without having to shard computation among multiple GPUs. We compare the computational efficiency of MLPs with contemporary networks of similar size such as \textit{ResNet-152}, \textit{ViT-B/4} and \textit{ViT-B/8} 
in Appendix \ref{sec:comp-efficiency}.

\section{Related Works}
 There are some prior works that investigate MLPs on vision tasks. \citet{Lin2015HowFC} study the performance of MLPs on small scale datasets such as CIFAR10. They observe similar improvements when using inverted bottleneck layers but do not study larger-scale setups, transfer-learning nor do they discuss the implications for theoretical works. The bottleneck structure used in this work has also been investigated theoretically \citep{Parhi2021WhatKO, Shenouda2023VectorValuedVS, parkinson2023linear}, further highlighting that such an architecture exhibits desirable properties. \citet{urban2017do} study to what degree convolutions are necessary for good performance and conclude that even with distillation techniques it remains very difficult to train performant MLPs on CIFAR10. Other approaches have focused on sparsifying fully-connected layers through evolutionary training \citep{evolutionary, 10.1145/2908812.2908890}, aiming to learn a good inductive bias from scratch. Similarly, \citet{NEURIPS2020_5c528e25} study how the inductive bias of MLPs can be improved by systematically sparsifying them with a LASSO-type algorithm, making them more convolution-like. \citet{NEURIPS2019_124c3e4a} on the other hand first train a convolutional network for a certain duration and then subsequently continue training the network as an MLP (by using the correspondence between CNNs and MLPs highlighted in Sec.~\ref{sec:background}). They show that good performance can be reached if the network was trained long enough as a CNN. In contrast to these works, our goal is not to enhance the inherent inductive bias of MLPs but study whether it can be overcome with enough scale. \\[2mm] The advent of the \textit{MLP-Mixer} \citep{tolstikhin2021mlpmixer} has led to a series of follow-up work, similarly using MLPs as a patch processor and token mixer \citep{touvron2021resmlp, chen2022cyclemlp, lian2022asmlp, Guo2021HireMLPVM, liu2021pay}. Again, we remark that these architectures all possess significantly more inductive bias. \\[2mm]
 Finally, we would like to remark that MLPs are successfully used in other areas such as novel view synthesis (e.g. NeRF \citep{nerf}).

\section{Discussion}
In this work, we have explored the limits of the multi-layer perceptron as an architecture for vision tasks. Our study reveals that (1) lack of inductive bias can be compensated by scale and (2) MLPs constitute a (largely) accurate proxy for modern architectures, further cementing their role as the main theoretical object of study. The role of data augmentation and the implicit bias of SGD however strongly differ for MLPs in the setting considered in this work and theoretical works should take this into account. Large-scale pre-training of MLPs proves to be very efficient, enabling researchers with less access to computational resources to study this very exciting line of work. While lack of inductive bias does not prevent MLPs from reaching impressive performance, it leads to an interesting shift in compute-optimality towards more training examples. Subjecting MLPs to even larger amounts of compute similar to \citet{Zhai_2022_CVPR}, especially in the form of more training examples, remains as very interesting future work.  

\bibliographystyle{apalike}
\bibliography{main}

\begin{thebibliography}{}

\bibitem[Allen-Zhu et~al., 2019a]{NEURIPS2019_62dad6e2}
Allen-Zhu, Z., Li, Y., and Liang, Y. (2019a).
\newblock Learning and generalization in overparameterized neural networks,
  going beyond two layers.
\newblock In Wallach, H., Larochelle, H., Beygelzimer, A., d\textquotesingle
  Alch\'{e}-Buc, F., Fox, E., and Garnett, R., editors, {\em Advances in Neural
  Information Processing Systems}, volume~32. Curran Associates, Inc.

\bibitem[Allen-Zhu et~al., 2019b]{pmlr-v97-allen-zhu19a}
Allen-Zhu, Z., Li, Y., and Song, Z. (2019b).
\newblock A convergence theory for deep learning via over-parameterization.
\newblock In {\em International Conference on Machine Learning}, pages
  242--252. PMLR.

\bibitem[Amari, 1967]{4039068}
Amari, S. (1967).
\newblock A theory of adaptive pattern classifiers.
\newblock {\em IEEE Transactions on Electronic Computers}, EC-16(3):299--307.

\bibitem[Anagnostidis et~al., 2023]{anagnostidis2023the}
Anagnostidis, S., Bachmann, G., Noci, L., and Hofmann, T. (2023).
\newblock The curious case of benign memorization.
\newblock In {\em The Eleventh International Conference on Learning
  Representations}.

\bibitem[Arora et~al., 2018]{pmlr-v80-arora18a}
Arora, S., Cohen, N., and Hazan, E. (2018).
\newblock On the optimization of deep networks: Implicit acceleration by
  overparameterization.
\newblock In {\em International Conference on Machine Learning}, pages
  244--253. PMLR.

\bibitem[Arora et~al., 2019a]{Arora2019OnEC}
Arora, S., Du, S.~S., Hu, W., Li, Z., Salakhutdinov, R., and Wang, R. (2019a).
\newblock On exact computation with an infinitely wide neural net.
\newblock In {\em Neural Information Processing Systems}.

\bibitem[Arora et~al., 2019b]{Arora2019FineGrainedAO}
Arora, S., Du, S.~S., Hu, W., Li, Z., and Wang, R. (2019b).
\newblock Fine-grained analysis of optimization and generalization for
  overparameterized two-layer neural networks.
\newblock In {\em International Conference on Machine Learning}.

\bibitem[Ba et~al., 2016]{ba2016layer}
Ba, J.~L., Kiros, J.~R., and Hinton, G.~E. (2016).
\newblock Layer normalization.

\bibitem[Bahri et~al., 2021]{bahri2021explaining}
Bahri, Y., Dyer, E., Kaplan, J., Lee, J., and Sharma, U. (2021).
\newblock Explaining neural scaling laws.

\bibitem[Battaglia et~al., 2018]{battaglia2018relational}
Battaglia, P.~W., Hamrick, J.~B., Bapst, V., Sanchez-Gonzalez, A., Zambaldi,
  V., Malinowski, M., Tacchetti, A., Raposo, D., Santoro, A., Faulkner, R.,
  Gulcehre, C., Song, F., Ballard, A., Gilmer, J., Dahl, G., Vaswani, A.,
  Allen, K., Nash, C., Langston, V., Dyer, C., Heess, N., Wierstra, D., Kohli,
  P., Botvinick, M., Vinyals, O., Li, Y., and Pascanu, R. (2018).
\newblock Relational inductive biases, deep learning, and graph networks.

\bibitem[Beyer et~al., 2020]{beyer2020imagenet}
Beyer, L., Hénaff, O.~J., Kolesnikov, A., Zhai, X., and van~den Oord, A.
  (2020).
\newblock Are we done with imagenet?

\bibitem[Brutzkus and Globerson, 2017]{pmlr-v70-brutzkus17a}
Brutzkus, A. and Globerson, A. (2017).
\newblock Globally optimal gradient descent for a {C}onv{N}et with {G}aussian
  inputs.
\newblock In Precup, D. and Teh, Y.~W., editors, {\em Proceedings of the 34th
  International Conference on Machine Learning}, volume~70 of {\em Proceedings
  of Machine Learning Research}, pages 605--614. PMLR.

\bibitem[Caballero et~al., 2023]{caballero2023broken}
Caballero, E., Gupta, K., Rish, I., and Krueger, D. (2023).
\newblock Broken neural scaling laws.
\newblock In {\em The Eleventh International Conference on Learning
  Representations}.

\bibitem[Caron et~al., 2021]{Caron2021EmergingPI}
Caron, M., Touvron, H., Misra, I., J'egou, H., Mairal, J., Bojanowski, P., and
  Joulin, A. (2021).
\newblock Emerging properties in self-supervised vision transformers.
\newblock {\em 2021 IEEE/CVF International Conference on Computer Vision
  (ICCV)}, pages 9630--9640.

\bibitem[Chen et~al., 2022]{chen2022cyclemlp}
Chen, S., Xie, E., GE, C., Chen, R., Liang, D., and Luo, P. (2022).
\newblock Cycle{MLP}: A {MLP}-like architecture for dense prediction.
\newblock In {\em International Conference on Learning Representations}.

\bibitem[Chen et~al., 2020]{pmlr-v119-chen20j}
Chen, T., Kornblith, S., Norouzi, M., and Hinton, G. (2020).
\newblock A simple framework for contrastive learning of visual
  representations.
\newblock In III, H.~D. and Singh, A., editors, {\em Proceedings of the 37th
  International Conference on Machine Learning}, volume 119 of {\em Proceedings
  of Machine Learning Research}, pages 1597--1607. PMLR.

\bibitem[Chen et~al., 2023]{chen2023symbolic}
Chen, X., Liang, C., Huang, D., Real, E., Wang, K., Liu, Y., Pham, H., Dong,
  X., Luong, T., Hsieh, C.-J., Lu, Y., and Le, Q.~V. (2023).
\newblock Symbolic discovery of optimization algorithms.

\bibitem[Chizat and Bach, 2020]{pmlr-v125-chizat20a}
Chizat, L. and Bach, F. (2020).
\newblock Implicit bias of gradient descent for wide two-layer neural networks
  trained with the logistic loss.
\newblock In Abernethy, J. and Agarwal, S., editors, {\em Proceedings of Thirty
  Third Conference on Learning Theory}, volume 125 of {\em Proceedings of
  Machine Learning Research}, pages 1305--1338. PMLR.

\bibitem[Chrabaszcz et~al., 2017]{chrabaszcz2017downsampled}
Chrabaszcz, P., Loshchilov, I., and Hutter, F. (2017).
\newblock A downsampled variant of imagenet as an alternative to the cifar
  datasets.

\bibitem[Coates et~al., 2011]{pmlr-v15-coates11a}
Coates, A., Ng, A., and Lee, H. (2011).
\newblock An analysis of single-layer networks in unsupervised feature
  learning.
\newblock In Gordon, G., Dunson, D., and Dudík, M., editors, {\em Proceedings
  of the Fourteenth International Conference on Artificial Intelligence and
  Statistics}, volume~15 of {\em Proceedings of Machine Learning Research},
  pages 215--223, Fort Lauderdale, FL, USA. PMLR.

\bibitem[Deng et~al., 2009]{5206848}
Deng, J., Dong, W., Socher, R., Li, L.-J., Li, K., and Fei-Fei, L. (2009).
\newblock Imagenet: A large-scale hierarchical image database.
\newblock In {\em 2009 IEEE Conference on Computer Vision and Pattern
  Recognition}, pages 248--255.

\bibitem[Dosovitskiy et~al., 2021]{dosovitskiy2021an}
Dosovitskiy, A., Beyer, L., Kolesnikov, A., Weissenborn, D., Zhai, X.,
  Unterthiner, T., Dehghani, M., Minderer, M., Heigold, G., Gelly, S.,
  Uszkoreit, J., and Houlsby, N. (2021).
\newblock An image is worth 16x16 words: Transformers for image recognition at
  scale.
\newblock In {\em International Conference on Learning Representations}.

\bibitem[d\textquotesingle Ascoli et~al., 2019]{NEURIPS2019_124c3e4a}
d\textquotesingle Ascoli, S., Sagun, L., Biroli, G., and Bruna, J. (2019).
\newblock Finding the needle in the haystack with convolutions: on the benefits
  of architectural bias.
\newblock In Wallach, H., Larochelle, H., Beygelzimer, A., d\textquotesingle
  Alch\'{e}-Buc, F., Fox, E., and Garnett, R., editors, {\em Advances in Neural
  Information Processing Systems}, volume~32. Curran Associates, Inc.

\bibitem[Du et~al., 2019]{du2018gradient}
Du, S.~S., Zhai, X., Poczos, B., and Singh, A. (2019).
\newblock Gradient descent provably optimizes over-parameterized neural
  networks.
\newblock In {\em International Conference on Learning Representations}.

\bibitem[Fernando et~al., 2016]{10.1145/2908812.2908890}
Fernando, C., Banarse, D., Reynolds, M., Besse, F., Pfau, D., Jaderberg, M.,
  Lanctot, M., and Wierstra, D. (2016).
\newblock Convolution by evolution: Differentiable pattern producing networks.
\newblock In {\em Proceedings of the Genetic and Evolutionary Computation
  Conference 2016}, GECCO '16, page 109–116, New York, NY, USA. Association
  for Computing Machinery.

\bibitem[Goyal et~al., 2017]{imagenet1hour}
Goyal, P., Dollár, P., Girshick, R., Noordhuis, P., Wesolowski, L., Kyrola,
  A., Tulloch, A., Jia, Y., and He, K. (2017).
\newblock Accurate, large minibatch sgd: Training imagenet in 1 hour.

\bibitem[Grill et~al., 2020]{NEURIPS2020_f3ada80d}
Grill, J.-B., Strub, F., Altch\'{e}, F., Tallec, C., Richemond, P.,
  Buchatskaya, E., Doersch, C., Avila~Pires, B., Guo, Z., Gheshlaghi~Azar, M.,
  Piot, B., kavukcuoglu, k., Munos, R., and Valko, M. (2020).
\newblock Bootstrap your own latent - a new approach to self-supervised
  learning.
\newblock In Larochelle, H., Ranzato, M., Hadsell, R., Balcan, M., and Lin, H.,
  editors, {\em Advances in Neural Information Processing Systems}, volume~33,
  pages 21271--21284. Curran Associates, Inc.

\bibitem[Gunasekar et~al., 2018]{NEURIPS2018_0e98aeeb}
Gunasekar, S., Lee, J.~D., Soudry, D., and Srebro, N. (2018).
\newblock Implicit bias of gradient descent on linear convolutional networks.
\newblock In Bengio, S., Wallach, H., Larochelle, H., Grauman, K.,
  Cesa-Bianchi, N., and Garnett, R., editors, {\em Advances in Neural
  Information Processing Systems}, volume~31. Curran Associates, Inc.

\bibitem[Guo et~al., 2021]{Guo2021HireMLPVM}
Guo, J., Tang, Y., Han, K., Chen, X., Wu, H., Xu, C., Xu, C., and Wang, Y.
  (2021).
\newblock Hire-mlp: Vision mlp via hierarchical rearrangement.
\newblock {\em 2022 IEEE/CVF Conference on Computer Vision and Pattern
  Recognition (CVPR)}, pages 816--826.

\bibitem[He et~al., 2015]{He2015DeepRL}
He, K., Zhang, X., Ren, S., and Sun, J. (2015).
\newblock Deep residual learning for image recognition.
\newblock {\em 2016 IEEE Conference on Computer Vision and Pattern Recognition
  (CVPR)}, pages 770--778.

\bibitem[Hestness et~al., 2019]{hestness2019humanlevel}
Hestness, J., Ardalani, N., and Diamos, G. (2019).
\newblock Beyond human-level accuracy: Computational challenges in deep
  learning.

\bibitem[Hestness et~al., 2017]{hestness2017deep}
Hestness, J., Narang, S., Ardalani, N., Diamos, G., Jun, H., Kianinejad, H.,
  Patwary, M. M.~A., Yang, Y., and Zhou, Y. (2017).
\newblock Deep learning scaling is predictable, empirically.

\bibitem[Hoffer et~al., 2017]{NIPS2017_a5e0ff62}
Hoffer, E., Hubara, I., and Soudry, D. (2017).
\newblock Train longer, generalize better: closing the generalization gap in
  large batch training of neural networks.
\newblock In Guyon, I., Luxburg, U.~V., Bengio, S., Wallach, H., Fergus, R.,
  Vishwanathan, S., and Garnett, R., editors, {\em Advances in Neural
  Information Processing Systems}, volume~30. Curran Associates, Inc.

\bibitem[Hoffmann et~al., 2022]{hoffmann2022training}
Hoffmann, J., Borgeaud, S., Mensch, A., Buchatskaya, E., Cai, T., Rutherford,
  E., de~Las~Casas, D., Hendricks, L.~A., Welbl, J., Clark, A., Hennigan, T.,
  Noland, E., Millican, K., van~den Driessche, G., Damoc, B., Guy, A.,
  Osindero, S., Simonyan, K., Elsen, E., Rae, J.~W., Vinyals, O., and Sifre, L.
  (2022).
\newblock Training compute-optimal large language models.

\bibitem[Hron et~al., 2020]{pmlr-v119-hron20a}
Hron, J., Bahri, Y., Sohl-Dickstein, J., and Novak, R. (2020).
\newblock Infinite attention: {NNGP} and {NTK} for deep attention networks.
\newblock In III, H.~D. and Singh, A., editors, {\em Proceedings of the 37th
  International Conference on Machine Learning}, volume 119 of {\em Proceedings
  of Machine Learning Research}, pages 4376--4386. PMLR.

\bibitem[Ivakhnenko et~al., 1965]{ivakhnenko1965cybernetic}
Ivakhnenko, A., Lapa, V., and ENGINEERING., P. U. L. I. S. O.~E. (1965).
\newblock {\em Cybernetic Predicting Devices}.
\newblock JPRS 37, 803. Joint Publications Research Service [available from the
  Clearinghouse for Federal Scientific and Technical Information].

\bibitem[Jacot et~al., 2018]{NEURIPS2018_5a4be1fa}
Jacot, A., Gabriel, F., and Hongler, C. (2018).
\newblock Neural tangent kernel: Convergence and generalization in neural
  networks.
\newblock In Bengio, S., Wallach, H., Larochelle, H., Grauman, K.,
  Cesa-Bianchi, N., and Garnett, R., editors, {\em Advances in Neural
  Information Processing Systems}, volume~31. Curran Associates, Inc.

\bibitem[Kaplan et~al., 2020]{kaplan2020scaling}
Kaplan, J., McCandlish, S., Henighan, T., Brown, T.~B., Chess, B., Child, R.,
  Gray, S., Radford, A., Wu, J., and Amodei, D. (2020).
\newblock Scaling laws for neural language models.

\bibitem[Keskar et~al., 2017]{keskar2017on}
Keskar, N.~S., Mudigere, D., Nocedal, J., Smelyanskiy, M., and Tang, P. T.~P.
  (2017).
\newblock On large-batch training for deep learning: Generalization gap and
  sharp minima.
\newblock In {\em International Conference on Learning Representations}.

\bibitem[Krizhevsky, 2009]{Krizhevsky2009LearningML}
Krizhevsky, A. (2009).
\newblock Learning multiple layers of features from tiny images.

\bibitem[Krizhevsky et~al., 2012]{NIPS2012_c399862d}
Krizhevsky, A., Sutskever, I., and Hinton, G.~E. (2012).
\newblock Imagenet classification with deep convolutional neural networks.
\newblock In Pereira, F., Burges, C., Bottou, L., and Weinberger, K., editors,
  {\em Advances in Neural Information Processing Systems}, volume~25. Curran
  Associates, Inc.

\bibitem[Le and Yang, 2015]{Le2015TinyIV}
Le, Y. and Yang, X.~S. (2015).
\newblock Tiny imagenet visual recognition challenge.

\bibitem[Leclerc et~al., 2023]{leclerc2023ffcv}
Leclerc, G., Ilyas, A., Engstrom, L., Park, S.~M., Salman, H., and Madry, A.
  (2023).
\newblock {FFCV}: Accelerating training by removing data bottlenecks.

\bibitem[Li and Yuan, 2017]{NIPS2017_a96b65a7}
Li, Y. and Yuan, Y. (2017).
\newblock Convergence analysis of two-layer neural networks with relu
  activation.
\newblock In Guyon, I., Luxburg, U.~V., Bengio, S., Wallach, H., Fergus, R.,
  Vishwanathan, S., and Garnett, R., editors, {\em Advances in Neural
  Information Processing Systems}, volume~30. Curran Associates, Inc.

\bibitem[Lian et~al., 2022]{lian2022asmlp}
Lian, D., Yu, Z., Sun, X., and Gao, S. (2022).
\newblock {AS}-{MLP}: An axial shifted {MLP} architecture for vision.
\newblock In {\em International Conference on Learning Representations}.

\bibitem[Lin et~al., 2015]{Lin2015HowFC}
Lin, Z., Memisevic, R., and Konda, K.~R. (2015).
\newblock How far can we go without convolution: Improving fully-connected
  networks.
\newblock {\em ArXiv}, abs/1511.02580.

\bibitem[Liu et~al., 2021]{liu2021pay}
Liu, H., Dai, Z., So, D., and Le, Q.~V. (2021).
\newblock Pay attention to {MLP}s.
\newblock In Beygelzimer, A., Dauphin, Y., Liang, P., and Vaughan, J.~W.,
  editors, {\em Advances in Neural Information Processing Systems}.

\bibitem[Liu et~al., 2022]{liu2022convnet}
Liu, Z., Mao, H., Wu, C.-Y., Feichtenhofer, C., Darrell, T., and Xie, S.
  (2022).
\newblock A convnet for the 2020s.
\newblock In {\em Proceedings of the IEEE/CVF Conference on Computer Vision and
  Pattern Recognition}, pages 11976--11986.

\bibitem[Maloney et~al., 2022]{maloney2022solvable}
Maloney, A., Roberts, D.~A., and Sully, J. (2022).
\newblock A solvable model of neural scaling laws.

\bibitem[Mei and Montanari, 2021]{mei2021generalization}
Mei, S. and Montanari, A. (2021).
\newblock The generalization error of random features regression: Precise
  asymptotics and the double descent curve.
\newblock {\em Communications on Pure and Applied Mathematics}, 75.

\bibitem[Mei et~al., 2018]{doi:10.1073/pnas.1806579115}
Mei, S., Montanari, A., and Nguyen, P.-M. (2018).
\newblock A mean field view of the landscape of two-layer neural networks.
\newblock {\em Proceedings of the National Academy of Sciences},
  115(33):E7665--E7671.

\bibitem[Mildenhall et~al., 2021]{nerf}
Mildenhall, B., Srinivasan, P.~P., Tancik, M., Barron, J.~T., Ramamoorthi, R.,
  and Ng, R. (2021).
\newblock Nerf: Representing scenes as neural radiance fields for view
  synthesis.
\newblock {\em Commun. ACM}, 65(1):99–106.

\bibitem[Mocanu et~al., 2018]{evolutionary}
Mocanu, D., Mocanu, E., Stone, P., Nguyen, P., Gibescu, M., and Liotta, A.
  (2018).
\newblock Scalable training of artificial neural networks with adaptive sparse
  connectivity inspired by network science.
\newblock {\em Nature Communications}, 9.

\bibitem[Neyshabur, 2020]{NEURIPS2020_5c528e25}
Neyshabur, B. (2020).
\newblock Towards learning convolutions from scratch.
\newblock In Larochelle, H., Ranzato, M., Hadsell, R., Balcan, M., and Lin, H.,
  editors, {\em Advances in Neural Information Processing Systems}, volume~33,
  pages 8078--8088. Curran Associates, Inc.

\bibitem[Neyshabur et~al., 2019]{neyshabur2018the}
Neyshabur, B., Li, Z., Bhojanapalli, S., LeCun, Y., and Srebro, N. (2019).
\newblock The role of over-parametrization in generalization of neural
  networks.
\newblock In {\em International Conference on Learning Representations}.

\bibitem[Neyshabur et~al., 2014]{neyshanur2014implicit}
Neyshabur, B., Tomioka, R., and Srebro, N. (2014).
\newblock In search of the real inductive bias: On the role of implicit
  regularization in deep learning.

\bibitem[OpenAI, 2018]{OPENAI_2018}
OpenAI (2018).
\newblock Ai and compute.

\bibitem[OpenAI, 2023]{openai2023gpt4}
OpenAI (2023).
\newblock Gpt-4 technical report.

\bibitem[Parhi and Nowak, 2021]{Parhi2021WhatKO}
Parhi, R. and Nowak, R.~D. (2021).
\newblock What kinds of functions do deep neural networks learn? insights from
  variational spline theory.
\newblock {\em SIAM J. Math. Data Sci.}, 4:464--489.

\bibitem[Parkinson et~al., 2023]{parkinson2023linear}
Parkinson, S., Ongie, G., and Willett, R. (2023).
\newblock Linear neural network layers promote learning single- and
  multiple-index models.

\bibitem[Paszke et~al., 2019]{NEURIPS2019_9015}
Paszke, A., Gross, S., Massa, F., Lerer, A., Bradbury, J., Chanan, G., Killeen,
  T., Lin, Z., Gimelshein, N., Antiga, L., Desmaison, A., Kopf, A., Yang, E.,
  DeVito, Z., Raison, M., Tejani, A., Chilamkurthy, S., Steiner, B., Fang, L.,
  Bai, J., and Chintala, S. (2019).
\newblock Pytorch: An imperative style, high-performance deep learning library.
\newblock In {\em Advances in Neural Information Processing Systems 32}, pages
  8024--8035. Curran Associates, Inc.

\bibitem[Poole et~al., 2016]{NIPS2016_14851003}
Poole, B., Lahiri, S., Raghu, M., Sohl-Dickstein, J., and Ganguli, S. (2016).
\newblock Exponential expressivity in deep neural networks through transient
  chaos.
\newblock In Lee, D., Sugiyama, M., Luxburg, U., Guyon, I., and Garnett, R.,
  editors, {\em Advances in Neural Information Processing Systems}, volume~29.
  Curran Associates, Inc.

\bibitem[Ridnik et~al., 2021]{imagenetformasses}
Ridnik, T., Ben-Baruch, E., Noy, A., and Zelnik, L. (2021).
\newblock Imagenet-21k pretraining for the masses.
\newblock In Vanschoren, J. and Yeung, S., editors, {\em Proceedings of the
  Neural Information Processing Systems Track on Datasets and Benchmarks},
  volume~1. Curran.

\bibitem[Roberts et~al., 2022]{roberts_yaida_hanin_2022}
Roberts, D.~A., Yaida, S., and Hanin, B. (2022).
\newblock {\em The Principles of Deep Learning Theory: An Effective Theory
  Approach to Understanding Neural Networks}.
\newblock Cambridge University Press.

\bibitem[Rosenblatt, 1958]{rosenblatt1958perceptron}
Rosenblatt, F. (1958).
\newblock The perceptron: a probabilistic model for information storage and
  organization in the brain.
\newblock {\em Psychological review}, 65(6):386.

\bibitem[Rosenfeld et~al., 2020]{rosenfeld2020a}
Rosenfeld, J.~S., Rosenfeld, A., Belinkov, Y., and Shavit, N. (2020).
\newblock A constructive prediction of the generalization error across scales.
\newblock In {\em International Conference on Learning Representations}.

\bibitem[Saxe et~al., 2014]{saxe2014exact}
Saxe, A.~M., McClelland, J.~L., and Ganguli, S. (2014).
\newblock Exact solutions to the nonlinear dynamics of learning in deep linear
  neural networks.

\bibitem[Schoenholz et~al., 2017]{schoenholz2017deep}
Schoenholz, S.~S., Gilmer, J., Ganguli, S., and Sohl-Dickstein, J. (2017).
\newblock Deep information propagation.
\newblock In {\em International Conference on Learning Representations}.

\bibitem[Shenouda et~al., 2023]{Shenouda2023VectorValuedVS}
Shenouda, J., Parhi, R., Lee, K., and Nowak, R.~D. (2023).
\newblock Vector-valued variation spaces and width bounds for dnns: Insights on
  weight decay regularization.
\newblock {\em ArXiv}, abs/2305.16534.

\bibitem[Soudry et~al., 2018]{soudry2018the}
Soudry, D., Hoffer, E., and Srebro, N. (2018).
\newblock The implicit bias of gradient descent on separable data.
\newblock In {\em International Conference on Learning Representations}.

\bibitem[Tan and Le, 2020]{tan2020efficientnet}
Tan, M. and Le, Q.~V. (2020).
\newblock Efficientnet: Rethinking model scaling for convolutional neural
  networks.

\bibitem[Tolstikhin et~al., 2021]{tolstikhin2021mlpmixer}
Tolstikhin, I., Houlsby, N., Kolesnikov, A., Beyer, L., Zhai, X., Unterthiner,
  T., Yung, J., Steiner, A.~P., Keysers, D., Uszkoreit, J., Lucic, M., and
  Dosovitskiy, A. (2021).
\newblock {MLP}-mixer: An all-{MLP} architecture for vision.
\newblock In Beygelzimer, A., Dauphin, Y., Liang, P., and Vaughan, J.~W.,
  editors, {\em Advances in Neural Information Processing Systems}.

\bibitem[Touvron et~al., 2021]{touvron2021resmlp}
Touvron, H., Bojanowski, P., Caron, M., Cord, M., El-Nouby, A., Grave, E.,
  Izacard, G., Joulin, A., Synnaeve, G., Verbeek, J., and Jegou, H. (2021).
\newblock Res{MLP}: Feedforward networks for image classification with
  data-efficient training.

\bibitem[Touvron et~al., 2023]{touvron2023llama}
Touvron, H., Lavril, T., Izacard, G., Martinet, X., Lachaux, M.-A., Lacroix,
  T., Rozi{\`e}re, B., Goyal, N., Hambro, E., Azhar, F., et~al. (2023).
\newblock Llama: Open and efficient foundation language models.
\newblock {\em arXiv preprint arXiv:2302.13971}.

\bibitem[Trockman and Kolter, 2022]{trockman2022patches}
Trockman, A. and Kolter, J.~Z. (2022).
\newblock Patches are all you need?

\bibitem[Urban et~al., 2017]{urban2017do}
Urban, G., Geras, K.~J., Kahou, S.~E., Aslan, O., Wang, S., Mohamed, A.,
  Philipose, M., Richardson, M., and Caruana, R. (2017).
\newblock Do deep convolutional nets really need to be deep and convolutional?
\newblock In {\em International Conference on Learning Representations}.

\bibitem[Vaswani et~al., 2017]{NIPS2017_3f5ee243}
Vaswani, A., Shazeer, N., Parmar, N., Uszkoreit, J., Jones, L., Gomez, A.~N.,
  Kaiser, L.~u., and Polosukhin, I. (2017).
\newblock Attention is all you need.
\newblock In Guyon, I., Luxburg, U.~V., Bengio, S., Wallach, H., Fergus, R.,
  Vishwanathan, S., and Garnett, R., editors, {\em Advances in Neural
  Information Processing Systems}, volume~30. Curran Associates, Inc.

\bibitem[Virtanen et~al., 2020]{2020SciPy-NMeth}
Virtanen, P., Gommers, R., Oliphant, T.~E., Haberland, M., Reddy, T.,
  Cournapeau, D., Burovski, E., Peterson, P., Weckesser, W., Bright, J., {van
  der Walt}, S.~J., Brett, M., Wilson, J., Millman, K.~J., Mayorov, N., Nelson,
  A. R.~J., Jones, E., Kern, R., Larson, E., Carey, C.~J., Polat, {\.I}., Feng,
  Y., Moore, E.~W., {VanderPlas}, J., Laxalde, D., Perktold, J., Cimrman, R.,
  Henriksen, I., Quintero, E.~A., Harris, C.~R., Archibald, A.~M., Ribeiro,
  A.~H., Pedregosa, F., {van Mulbregt}, P., and {SciPy 1.0 Contributors}
  (2020).
\newblock {{SciPy} 1.0: Fundamental Algorithms for Scientific Computing in
  Python}.
\newblock {\em Nature Methods}, 17:261--272.

\bibitem[You et~al., 2017]{you2017large}
You, Y., Gitman, I., and Ginsburg, B. (2017).
\newblock Large batch training of convolutional networks.

\bibitem[Yu et~al., 2022]{yu2022metaformer}
Yu, W., Luo, M., Zhou, P., Si, C., Zhou, Y., Wang, X., Feng, J., and Yan, S.
  (2022).
\newblock Metaformer is actually what you need for vision.

\bibitem[Zhai et~al., 2022]{Zhai_2022_CVPR}
Zhai, X., Kolesnikov, A., Houlsby, N., and Beyer, L. (2022).
\newblock Scaling vision transformers.
\newblock In {\em Proceedings of the IEEE/CVF Conference on Computer Vision and
  Pattern Recognition (CVPR)}, pages 12104--12113.

\bibitem[Zhang et~al., 2018]{zhang2018mixup}
Zhang, H., Cisse, M., Dauphin, Y.~N., and Lopez-Paz, D. (2018).
\newblock mixup: Beyond empirical risk minimization.
\newblock In {\em International Conference on Learning Representations}.

\bibitem[Zou et~al., 2020]{zou2020gradient}
Zou, D., Cao, Y., Zhou, D., and Gu, Q. (2020).
\newblock Gradient descent optimizes over-parameterized deep relu networks.
\newblock {\em Machine Learning}, 109:1--26.

\end{thebibliography}

\newpage
\begin{appendices}

\textbf{\Large Appendix}
\section{Experimental Details}
\subsection{Resources}
For all experiments we rely on NVIDIA RTX A5000 GPU with 24GB of memory. Every experiment can be performed on a single GPU. We leverage the FFCV dataloader framework since the transfer time of the data to the GPU becomes the bottleneck in terms of training time in case of MLPs. All of our experiments were performed in PyTorch \citep{NEURIPS2019_9015}.

\subsection{Additional Ablations}
\begin{figure}[!ht]
    \centering
    \includegraphics[width=1.0\textwidth]{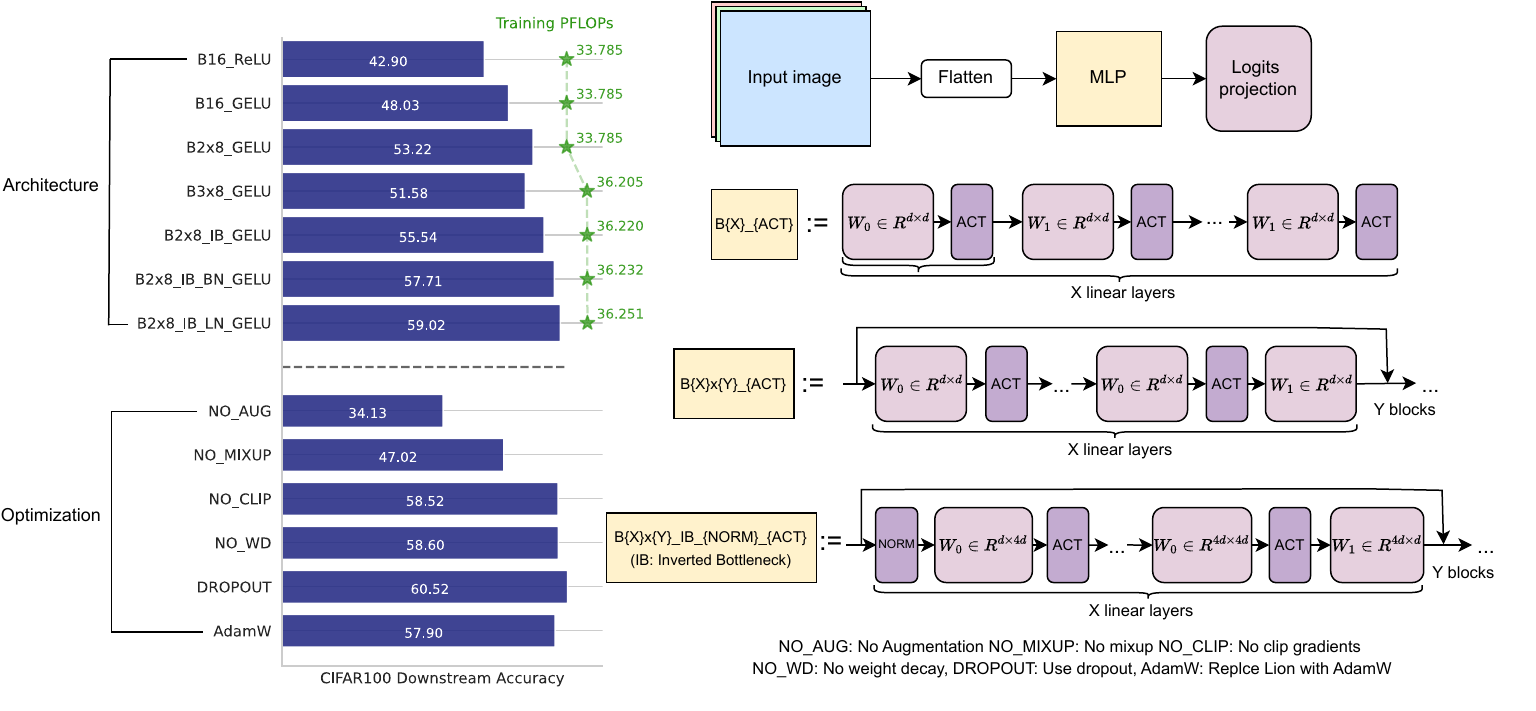}
    \caption{Ablations on different architectures and optimizations choices when training on ImageNet. Numbers indicate linear probing Top-1 accuracies on CIFAR100.}
    \label{fig:ablations}
\end{figure}

\paragraph{Ablations.} We provide some more ablations in Fig.~\ref{fig:ablations}. More specifically, for a (approximate) fixed budget of compute, we investigate different architecture and optimization choices, when pretraining on ImageNet1k and performing linear probing on CIFAR100. 
\label{sec:ablations}

\paragraph{Normalization.} We investigate the importance of the normalization method (LayerNorm vs BatchNorm) in more detail in Table \ref{tab:normalization}. We pre-train two B-MLPs on ImageNet21k with layer normalization and batch normalization and compare the fine-tuning performance on various tasks. We find that the techniques perform similarly, which layer normalisation having a slight edge.  
\begin{table}[!h]
\vskip 0.15in
    \centering
\begin{small}
\begin{sc}
\begin{tabular}{lccccc}
\toprule
 & CIFAR-10 & CIFAR-100  & TinyImageNet & ImageNet \\
\midrule
LayerNorm  & $90.0$ &  $74.6$ & $59.6$ & $36.2$ \\[1mm]
\midrule
BatchNorm  & $89.4$ &  $73.8$ & $57.7$ & $35.9$ \\[1mm]
\bottomrule
\end{tabular}
\end{sc}
\end{small}
\caption{Pretraining a B-6/Wi-1024 B-MLP with BatchNorm and LayerNorm on ImageNet21k and subsequently fine-tuning.}
    \label{tab:normalization}
\end{table}
\paragraph{Label smoothing.}
We further ablate the influence of label smoothing on the downstream performance. We pre-train B-MLPs with varying amounts of label smoothing ($\alpha \in \{0.0, 0.1, 0.3\}$) and evaluate the resulting down-stream fine-tuning performance. We report the results in Table \ref{tab:smoothing}. While label smoothing does provide some boost in performance, the gains are very modest. Label smoothing is thus helpful but not essential for training MLPs.
\begin{table}[]
    \centering
    \begin{small}
    \begin{sc}
    \begin{tabular}{lccccc}
\toprule
 & CIFAR10 & CIFAR100  & TinyImageNet & ImageNet \\
\midrule
$\alpha=0.3$  & $90.0$ &  $74.6$ & $59.6$ & $36.2$ \\[1mm]
\midrule
$\alpha=0.1$  & $89.5$ &  $73.7$ & $58.2$ & $36.0$ \\[1mm]
\midrule
$\alpha=0.0$  & $89.2$ &  $72.2$ & $57.1$ & $35.7$ \\[1mm]
\bottomrule
\end{tabular}
\end{sc}
\end{small}
    \caption{Pretraining a B-6/Wi-1024 B-MLP with different amounts of label smoothing on ImageNet21k and subsequently fine-tuning.}
    \label{tab:smoothing}
\end{table}

\paragraph{Architecture.}We make the following observations/recommendations to boost the model's performance, in line with results reported in the literature~\citep{liu2022convnet}; (1) replacing ReLUs and GELUs boosts results significantly, (2) adding skip connections every two layers helps with optimization, especially for deeper networks. (3) Using an inverted bottleneck increases performance even more. (4) Using a normalization layer in the PRE-LN configuration helps with optimization and (4) layer normalization leads to significantly better results compared to batch normalization, while also being more stable during training. 

\paragraph{Optimization.}As discussed in the main text, augmentations are crucial, and disabling them can have a detrimental effect. We also found that clipping gradients, using weight decay and dropout have a small positive effect on downstream performance. Finally, replacing LION~\citep{chen2023symbolic} with Adam(W), leads to a decrease in performance.

\subsection{Linear Probing}
\label{sec:linear-probing}
We showcase the transferability of our MLPs by training a linear classifier on top of the frozen features. For training the linear layer, we use the LION optimizer with a learning rate of $\eta = 0.00001$ for $50$ epochs. We display the results in Table \ref{tab:linear}.
\begin{table}[!ht]
\vskip 0.15in
\begin{center}
\begin{small}
\begin{sc}
\begin{tabular}{lccccc}
\toprule
 & CIFAR10 & CIFAR100 & STL10 & TinyImageNet & ImageNet \\
\midrule
\textit{{B}-6/{Wi}-1024}  & $65.1$ & $41.3$ & $53.4$  & $45.6$ & $13.0$ \\[1mm]
\midrule
\textit{{B}-6/{Wi}-1024} + DA  & $87.8$ & $73.2$ & $85.2$  & $61.3$ & $39.2$ \\[1mm]
\midrule
\textit{{B}-12/{Wi}-1024} + DA & $90.6$ & $74.5$ & $88.3$ & $68.5$ & $40.7$ \\[1mm]

\bottomrule
\end{tabular}
\end{sc}
\end{small}
\end{center}
\caption{Linear probing Top-1 accuracies when pretraining on ImageNet21k.}
\label{tab:linear}
\end{table}
We observe very strong down-stream performance even in this more limited setting, highlighting how transferable the features learnt by MLPs are.  

\subsection{Scaling Laws}
\label{sec: more-scaling-laws}
\paragraph{Implementation Details.} For the scaling law plots, we trained all the models with a batch-size $16384$ and the LION optimizer with a learning rate $\eta=0.00001$ and weight decay of strength $0.001$. We further use label smoothing of strength $0.3$. We again use augmentations in the form of random flips and crops as well as MixUp with strength $0.8$. We rely on the \textit{curvefit} function from the \textit{SciPy} library \citep{2020SciPy-NMeth} to fit powerlaws of the form $E(C) = a(b + C)^{-\alpha}+E_{\infty}$.

\subsection{Computational Efficiency}
We highlight the fact that although MLPs require a lot of training data, inference is extremely efficient from a computational perspective. To illustrate this, we embark on the following comparison; we study inference on $64 \times 64$ resolution images in an MLP vs other popular vision architectures of similar size and complexity in Table~\ref{tab:hardware_stats}. More specifically, we compare against a \textit{ResNet-152}, where we replace the stride in the first convolutional layer and remove the first max-pooling operation to compensate for the smaller image size. We also compare against a base \textit{ViT} and \textit{Mixer} model, where we extract patches from $4 \times 4$ regions in the original image. 

As it quickly becomes eminent, MLPs require significantly less FLOPs to make predictions on individual images, in essence utilizing their parameters a lot more methodically. As a result, latency and throughput are significantly better compared to other candidate architectures. We measure throughput using the optimal batch size on an NVIDIA RTX A5000. We highlight, that our MLPs, in contrast to the other architectures are memory bound, meaning that their throughput is determined by the prefetching bandwidth of our GPU. Hardware advancement and specialized architectures could significantly mitigate this effect. Neglecting memory transfer time by propagating the same input through our network gives a further $6$-fold increase in the potential throughput.

\label{sec:comp-efficiency}
\begin{table}[!h]
\vskip 0.15in
\begin{center}
\begin{small}
\begin{sc}
\begin{tabular}{lccccc}
\toprule
 & Parameters & \makecell{Latency \\ (msec)} & \makecell{Throughput \\ (images/sec)} & FLOPs per forward pass \\
\midrule
\textit{{B}-12/{Wi}-768}  & 66.89 \textit{M} & 21.2 & 16063 & 66.8 \textit{M} \\[1mm]
\midrule
\textit{ResNet-152} & 60.19 \textit{M} & 423 & 506 & 13.07 \textit{G} \\[1mm]
\midrule
\textit{ViT-B/4}  & 86.06 \textit{M} & 424 & 222 & 23.08 \textit{G} \\[1mm]
\midrule
\textit{Mixer-B/4}  & 63.82 \textit{M} & 400 & 319 & 19.36 \textit{G} \\[1mm]
\bottomrule
\end{tabular}
\end{sc}
\end{small}
\end{center}
\caption{Various measures assessing the computational efficiency of different architectures.}
\label{tab:hardware_stats}
\end{table}

\section{Results for Standard MLPs}
\subsection{Transfer Learning}
\label{sec:transfer-standard}
For completeness we also analyze the transfer performance of standard MLPs when pre-trained on ImageNet21k. We compare a S-MLP of depth $6$ and width $2048$ against a B-MLP of depth $6$ and width $1024$, ensuring that both models roughly have a parameter count of around $\approx70$ million. We display the results in Table \ref{tab:transfer-standard}. We observe that even the features of a standard MLP (i.e. without residual connections and bottleneck structure) transfer very well on different downstream task. Tthe inverted-bottleneck MLP however still remains superior. 
\begin{table}[!h]
    \centering
    \begin{small}
    \begin{sc}
    
    \begin{tabular}{lccccc}
    \toprule
     & CIFAR10 & CIFAR100  & TinyImageNet & ImageNet \\
    \midrule
    
    S-MLP & $87.1$ &  $68.3$ & $52.1$ & $30.2$ \\[1mm]
    \midrule
    B-MLP  & $90.0$ &  $74.6$ & $59.6$ & $36.2$ \\[1mm]
    \bottomrule
    \end{tabular}
    \end{sc}
    \end{small}
    \caption{Comparing a S-MLP of width 2048 and depth 6 pre-trained on ImageNet21k, with a B-6/Wi-1024 B-MLP (both models around 70M params) in terms of fine-tuning performance }
    \label{tab:transfer-standard}
\end{table}

\subsection{Scaling Laws}
\label{sec:scaling-standard}
We also evaluate the scaling law of standard MLPs by training variously sized models on different subsets of ImageNet21k and subsequently linearly probing the features on CIFAR100. The setting is identical to the one described in \ref{sec:scaling-laws}. We observe that also standard MLPs exhibit power-law behaviour. The slope ($0.22$ vs $0.25$) and the intercept ($0.18$ vs $0.16$) are however worse when compared against the inverted-bottleneck MLP.
\begin{figure}
    \centering
\includegraphics[width=0.6\textwidth]{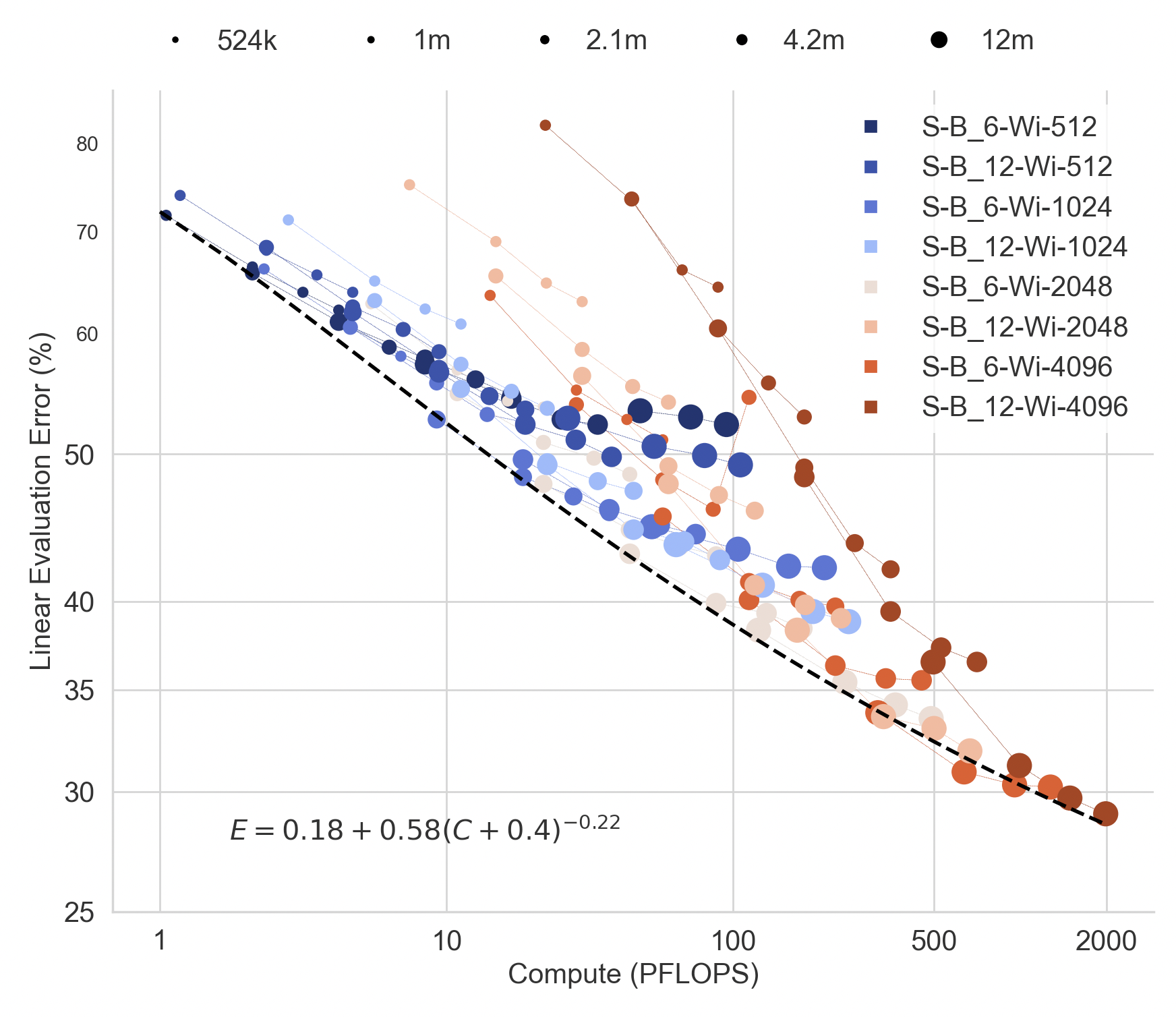}
    \caption{Test error of standard MLPs on CIFAR100 when linearly transferred as a function of PFLOPS, measured according to Eq.\eqref{eq:compute}, on a log-log scale.}
    \label{fig:enter-label}
\end{figure}

\section{Weight visualizations}
\begin{figure}
    \centering
    \includegraphics[width=0.9\textwidth]{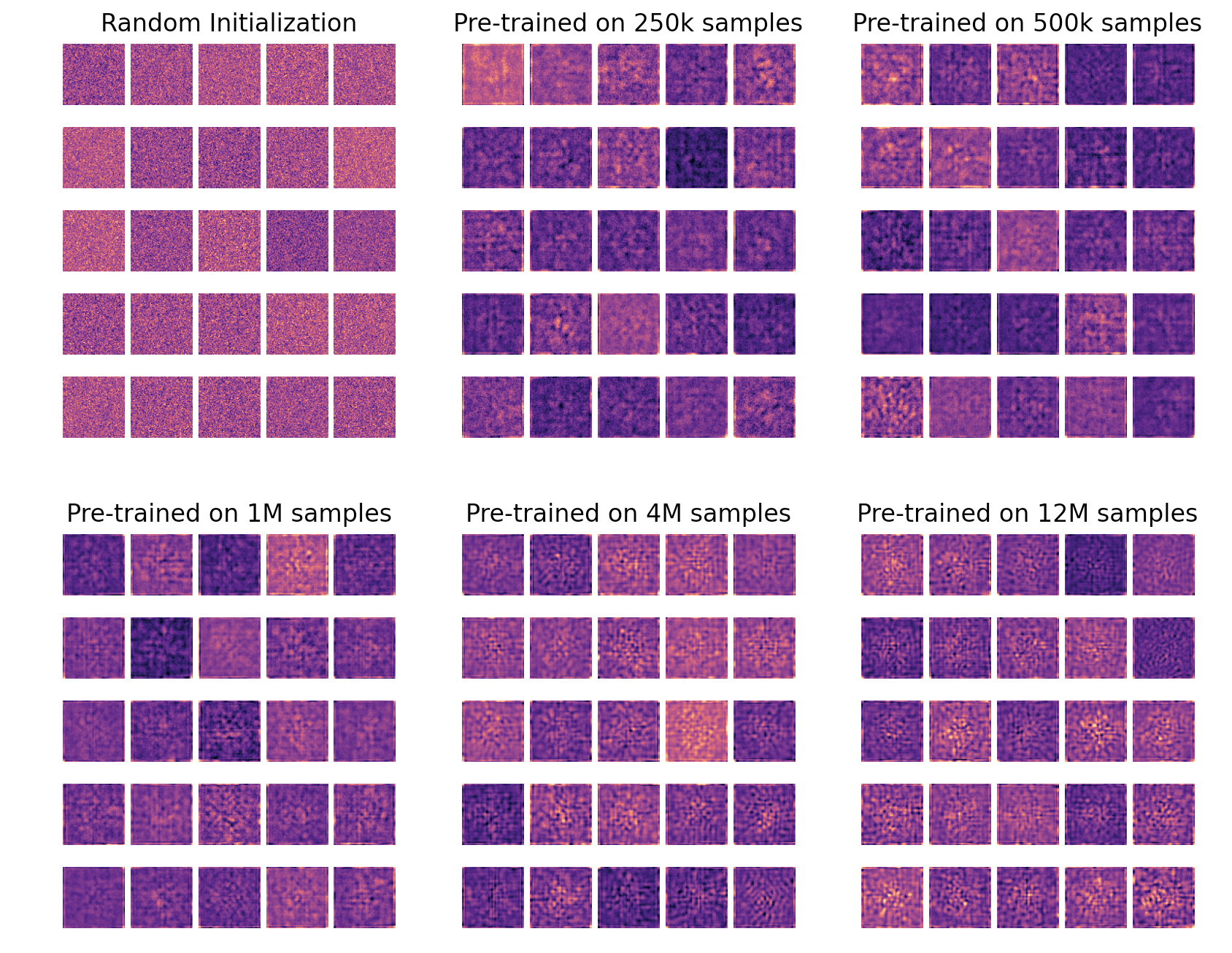}
    \caption{Visualization of the first layer weights for different pre-training dataset sizes.}
    \label{fig:visualization-pretraining}
\end{figure}

\begin{figure}
    \centering
    \includegraphics[width=0.9\textwidth]{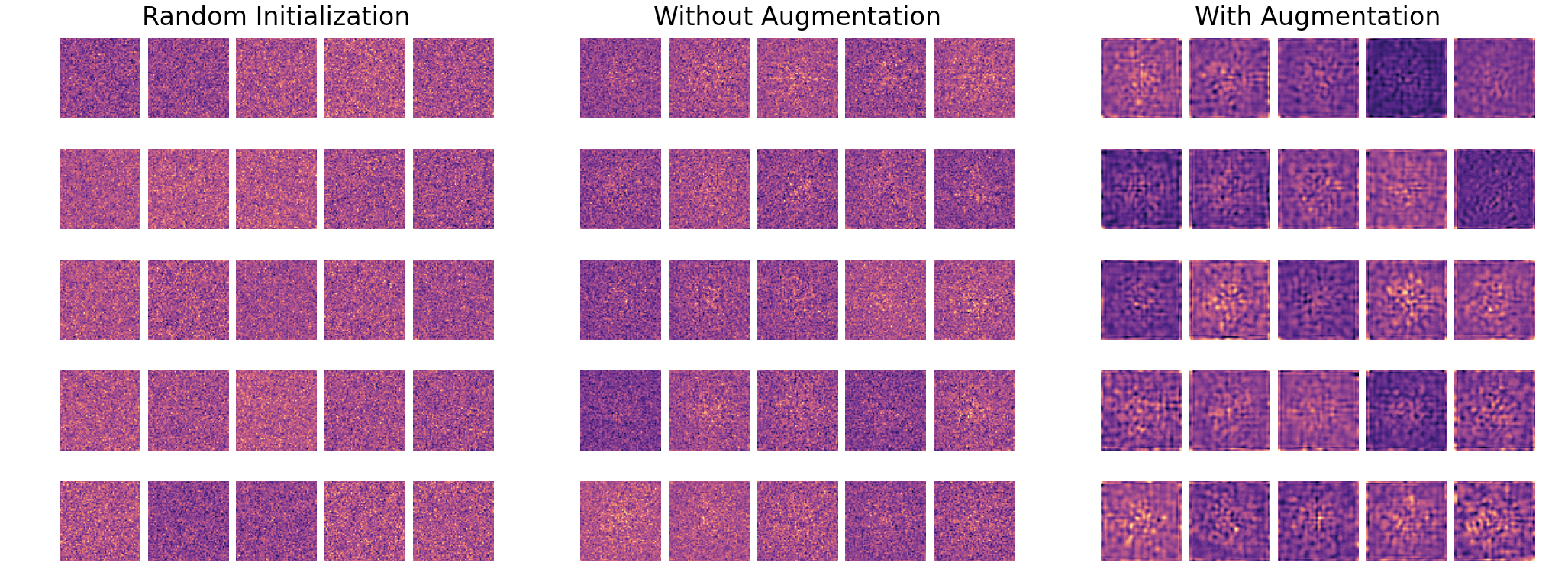}
    \caption{Visualization of the first layer weights for models trained with and without data augmentation.}
    \label{fig:visualization-aug}
\end{figure}

We visualize the first layer weights $\bm{W}^{(1)} \in \mathbb{R}^{3wh \times m}$ by reshaping them back to $\mathbb{R}^{w \times h \times 3 \times m}$. We then produce a $\mathbb{R}^{w \times h \times m}$ representation by taking the maximal value along the channel dimension. We display such visualizations of the first $5 \times 5 = 25$ "filters" for different pre-training sizes, also including the weights at random initialization in Fig.~\ref{fig:visualization-pretraining}. All models were trained with data augmentation. We observe that filters increasingly develop structure as we increase the dataset size and become more and more localized.
We further compare against models that were pre-trained on the full ImageNet21k, with and without data augmentation in Fig.~\ref{fig:visualization-aug}. We observe that even though we provide the model with an abundance of samples, the weights still remain largely structure-less and have not developped any locality properties. On the other hand, using data augmentation leads to more adapted filters.
\label{sec:visualizations-features}

\section{Inverted Bottleneck MLP Code}
We provide PyTorch-style pseudo-code for the inverted bottleneck MLP to highlight its simplicity. 
\begin{minipage}{\linewidth}
\begin{small}
\begin{lstlisting}[language=Python]
from torch import nn

class Block(nn.Module):
    def __init__(self, dim, expansion_factor=4, dropout=0.):
        super().__init__()
        self.fn = nn.Sequential(
            nn.Linear(dim, int(expansion_factor * dim)),
            nn.GELU(),
            nn.Dropout(dropout),
            nn.Linear(int(expansion_factor * dim), dim),
            nn.Dropout(dropout)
        )
        self.ln = nn.LayerNorm(dim)
        
    def forward(self, x):
        return x + self.fn(self.ln(x))
        

def MLP(image_size, channels, dim, depth, num_classes, expansion_factor=4, dropout=0.):
    return nn.Sequential(
        nn.Flatten(start_dim=1, end_dim=-1),
        nn.Linear(image_size * image_size * channels, dim),
        *[Block(dim, expansion_factor, dropout) for _ in range(depth)],
        nn.Linear(dim, num_classes)
    )
\end{lstlisting}
\end{small}
\end{minipage}


\end{appendices}
\end{document}